\documentclass{article} 
\usepackage{colm}
\usepackage[colorlinks = true,
            linkcolor = blue,
            urlcolor  = blue,
            citecolor = blue,
            anchorcolor = blue]{hyperref}   
\usepackage[utf8]{inputenc} 
\usepackage[T1]{fontenc}    
\usepackage{url}            
\usepackage{booktabs}       
\usepackage{amsfonts}       
\usepackage{nicefrac}       
\usepackage{microtype}      
\usepackage{xcolor}         

\usepackage{booktabs}
\usepackage{multirow}
\usepackage{tabularx}
\usepackage{makecell}
\usepackage{amssymb}
\usepackage{pifont}
\usepackage{amsmath}
\usepackage{tcolorbox}
\usepackage{listings}
\usepackage{xcolor}
\tcbuselibrary{skins}
\usepackage{hyperref}
\usepackage{float}
\usepackage{tcolorbox}
\tcbuselibrary{breakable}

\usepackage{fontawesome}

\usepackage{bbding}
\makeatletter
  \newcommand\figcaption{\def\@captype{figure}\caption}
  \newcommand\tabcaption{\def\@captype{table}\caption}
\makeatother

\usepackage[utf8]{inputenc} 
\usepackage[T1]{fontenc}    
\usepackage{url}            
\usepackage{booktabs}       
\usepackage{amsfonts}       
\usepackage{nicefrac}       
\usepackage{microtype}      
\usepackage{array}
\newcolumntype{C}[1]{>{\centering\arraybackslash}m{#1}}
\usepackage{xcolor}         
\usepackage{color, colortbl}
\definecolor{citecolor}{HTML}{2980b9}
\definecolor{linkcolor}{HTML}{c0392b}
\definecolor{darkorange}{HTML}{FF8C00}
\definecolor{chocolate}{HTML}{D2691E}
\definecolor{darkgreen}{HTML}{006400}
\definecolor{darkblue}{HTML}{00008B}
\definecolor{mediumblue}{HTML}{0000CD}
\definecolor{dodgerblue}{HTML}{1E90FF}
\definecolor{royalblue}{HTML}{4169E1}
\definecolor{shadecolor}{RGB}{237,237,237}
\definecolor{backred}{RGB}{255, 190, 190}
\definecolor{backblue}{RGB}{210, 230, 250}
\usepackage{colortbl}
\definecolor{graybg}{gray}{0.9}
\definecolor{zrrgreen}{HTML}{008000}
\definecolor{zrrblue}{HTML}{4682B4}
\definecolor{zrrred}{HTML}{B22222}
\definecolor{purple1}{RGB}{126, 107, 196}
\definecolor{purple2}{RGB}{199, 158, 207}
\definecolor{purple3}{RGB}{214, 200, 255}
\definecolor{purple4}{RGB}{254, 240, 255}
\usepackage{amsmath}
\usepackage{amssymb}
\usepackage{graphicx}

\usepackage{multirow}
\usepackage{makecell}
\usepackage{caption}

\usepackage{adjustbox}
\usepackage{wrapfig}

\usepackage{float}
\usepackage{subfig}
\usepackage[linesnumbered,ruled,vlined]{algorithm2e}


\usepackage{framed}
\usepackage{makecell}
\usepackage{listings}
\definecolor{lightgray}{rgb}{.9,.9,.9}
\definecolor{darkgray}{rgb}{.4,.4,.4}
\definecolor{purple}{rgb}{0.65, 0.12, 0.82}
\lstdefinelanguage{JavaScript}{
  keywords={break, case, catch, continue, debugger, default, delete, do, else, false, finally, for, function, if, in, instanceof, new, null, return, switch, this, throw, true, try, typeof, var, void, while, with},
  morecomment=[l]{//},
  morecomment=[s]{/*}{*/},
  morestring=[b]',
  morestring=[b]",
  ndkeywords={class, export, boolean, throw, implements, import, this},
  keywordstyle=\color{blue}\bfseries,
  ndkeywordstyle=\color{darkgray}\bfseries,
  identifierstyle=\color{black},
  commentstyle=\color{purple}\ttfamily,
  stringstyle=\color{red}\ttfamily,
  sensitive=true
}
\usepackage{xspace}
\usepackage[shortlabels]{enumitem}

\newcommand\blfootnote[1]{%
  \begingroup
  \renewcommand\thefootnote{}\footnote{#1}%
  \addtocounter{footnote}{-1}%
  \endgroup
}

\title{ArenaRL: Scaling RL for Open-Ended Agents via Tournament-based Relative Ranking}

\author{Qiang Zhang$^{1,\dagger}$, Boli Chen$^{1}$, Fanrui Zhang$^{1,\dagger}$, Ruixue Ding$^{1}$$^{\ast}$, Shihang Wang$^{1}$, Qiuchen Wang$^{1}$\\ \textbf{Yinfeng Huang$^{2}$, Haonan Zhang$^{2}$, Rongxiang Zhu$^{2}$, Pengyong Wang$^{2}$, Ailin Ren$^{2}$, Xin Li$^{2}$}\\ \textbf{Pengjun Xie$^{1}$, Jiawei Liu$^{\ddagger}$, Ning Guo$^{2}$, Jingren Zhou$^{1}$, Zheng-Jun Zha}\\
\\
$^{1}$\includegraphics[height=0.4cm]{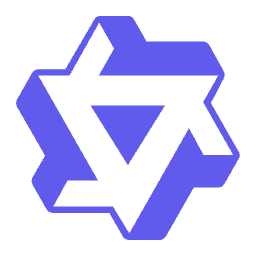} Tongyi Lab, Alibaba Group $^{2}$\includegraphics[height=0.4cm]{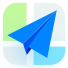} Amap, Alibaba Group
}

\colmfinalcopy 

\begin{document}

\maketitle

\blfootnote{$^{\dagger}$ Equal contribution, $^{\ast}$ Project leader, $^{\ddagger}$ Corresponding author}

\begin{abstract}
Reinforcement learning (RL) has substantially improved the performance of large language model (LLM) agents on tasks with verifiable outcomes, such as mathematics and code generation, but it still struggles on open-ended agent tasks with vast solution spaces (\textit{e.g.}, complex travel planning). Due to the absence of objective ground-truth for these tasks, current RL algorithms largely rely on reward models that assign scalar scores to individual responses. We contend that such pointwise scoring suffers from an inherent discrimination collapse: the reward model struggles to distinguish subtle advantages among different trajectories, resulting in scores within a group being compressed into a narrow range. Consequently, the effective reward signal becomes dominated by noise from the reward model, leading to optimization stagnation.
To tackle this issue, we propose ArenaRL, a reinforcement learning paradigm that shifts from pointwise scalar scoring to intra-group relative ranking. 
ArenaRL introduces a process-aware pairwise evaluation mechanism, employing multi-level rubrics to assign fine-grained relative scores to trajectories. Additionally, we construct an intra-group adversarial arena and devise a tournament-based ranking scheme to obtain stable advantage signals. Empirical results confirm that the built seeded single-elimination scheme achieves nearly equivalent advantage estimation accuracy to full pairwise comparisons with $\mathcal{O}(N^2)$ complexity, while operating with only $\mathcal{O}(N)$ complexity, striking an optimal balance between efficiency and precision.
Furthermore, to address the lack of full-cycle benchmarks for open-ended agents, we build Open-Travel and Open-DeepResearch, two high-quality benchmarks featuring a comprehensive pipeline covering supervised fine-tuning, RL training, and multi-dimensional automated evaluation. Extensive experiments show that ArenaRL substantially outperforms standard RL baselines, enabling LLM agents to generate solutions that are more logically rigorous and robust on complex real-world tasks.
The code is available at \href{https://github.com/Alibaba-NLP/qqr}{https://github.com/Alibaba-NLP/qqr}.

\end{abstract}

\section{Introduction}

The evolution of large language models (LLMs) into autonomous agents marks a paradigm shift in artificial intelligence from passive question answering to active problem solving. By integrating long-horizon planning and tool use, such agents have demonstrated substantial potential in handling complex tasks \cite{yao2022react, li2025deepagent, team2025tongyi}. In this progression, reinforcement learning (RL) has played a pivotal role, particularly in deterministic tasks such as mathematical reasoning and code generation, where ground-truth labels provide explicit reward signals for optimization \cite{dong2025agentic, li2025webthinker}. However, extending RL to open-ended agent tasks in real-world scenarios, such as personalized travel planning or in-depth industry analysis, poses fundamental challenges \cite{li2025webweaver}. In these domains, the solution spaces are vast and unstructured, and the notion of correctness is inherently subjective and multi-dimensional, rendering traditional ground-truth based RL inapplicable \cite{jia2025writing}.

\begin{figure}[t]
    \centering
    \includegraphics[width=1.0\textwidth]{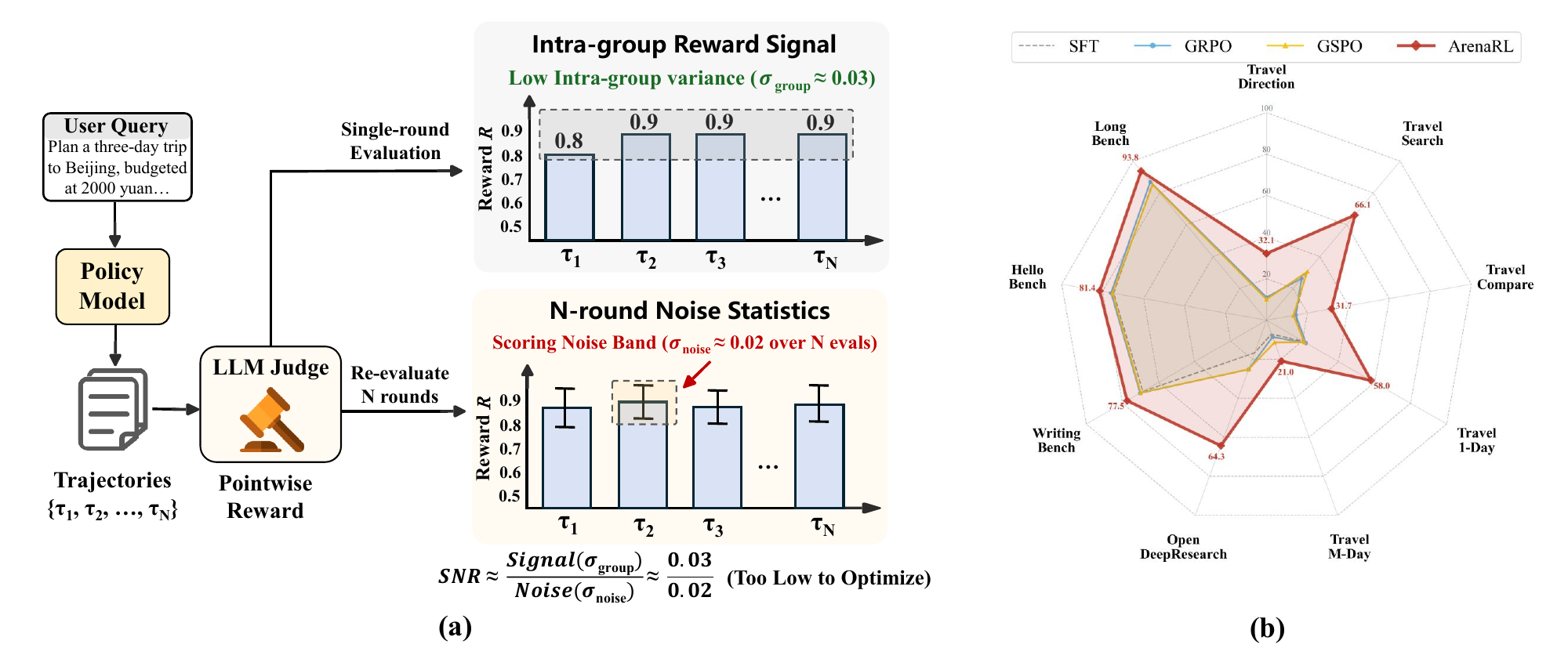}
    \caption{
    \textbf{(a) Illustration of discriminative collapse in pointwise evaluation:} We analyse the rewards of trajectory groups generated for a query example during RL training through two evaluation settings. First, the intra-group reward signal represents the pointwise rewards assigned to each trajectory within the group during a single-round evaluation. And the intra-group variance, denoted as $\sigma_{\text{group}}$, quantifies the degree of variation among different trajectories. Secondly, the N-round noise statistic present the average reward and corresponding scoring noise band for each trajectory across N independent evaluation repetitions, from which the noise variance $\sigma_{\text{noise}}$ is estimated. Observations reveal that the evaluation noise variance $\sigma_{\text{noise}}$ is substantial, comparable to the intra-group variance $\sigma_{\text{group}}$. This results in an extremely low signal-to-noise ratio (SNR), causing genuine advantages to be obscured by noise and hindering effective reinforcement learning optimization.
    \textbf{(b) ArenaRL Performance:} By shifting from pointwise scalar scoring to tournament-based relative ranking, ArenaRL significantly outperforms baselines (SFT, GRPO, GSPO) across diverse open-ended benchmarks.}
    \label{fig:first}
\end{figure}

Mainstream RL approaches (\textit{e.g.}, GRPO \cite{grpo}, GSPO \cite{gspo}) typically assume the existence of a verifiable reward function that can provide accurate reward feedback. However, this assumption often breaks down in open-ended settings where no objective ground truth is available. To address the issue of reward acquisition, recent work has adopted the LLM-as-Judge paradigm, assigning pointwise scalar scores to model outputs \cite{viswanathan2025checklists, huang2025reinforcement, liu2025openrubrics}. We identify that this mechanism leads to a severe phenomenon that we term discriminative collapse, as shown in Figure \ref{fig:first}(a). As the policy is progressively refined, the generated trajectories become increasingly similar in distribution. Consequently, rewards for trajectories within the same group are compressed by the pointwise scoring scheme into a narrow range (\textit{e.g.}, 0.8–0.9 on a scale of 1), rendering rewards indistinguishable. Moreover, due to inherent noise in the LLM judge, such as decoding randomness \cite{arias2025decoding} and length preferences \cite{hu2024explaining}, the reward outcomes exhibit a certain degree of unreliability, with a low signal-to-noise ratio between the reward signal and the interfering noise. In this situation, the pointwise evaluation mechanism struggles to distinguish truly superior samples. And the RL optimization process is driven more by spurious noise than by meaningful task-specific rewards, leading performance to stagnate or even deteriorate.

To fundamentally address discriminative collapse, we draw inspiration from decision theory, where pairwise preference judgments are known to be more stable than pointwise quantitative assessments \cite{furnkranz2010preference, rafailov2023direct}, and advocate a paradigm shift from pointwise scalar scoring to intra-group relative ranking. To this end, we propose ArenaRL, an online policy optimization framework grounded in an adversarial arena. ArenaRL abandons unstable scalar rewards in favor of constructing relative rankings over trajectories within each generated group. 
To ensure both depth and fairness in evaluation, we introduce a process-aware pairwise evaluation mechanism that not only compares the reliability of the outcome, but also scrutinizes the logical coherence of the chain-of-thought and the effectiveness of tool invocations along the trajectory.

A central bottleneck in scaling pairwise preference optimization to open-ended agent tasks lies in the high computational cost. While exhaustive comparisons yield accurate rankings, the resulting $\mathcal{O}(N^2)$ complexity is intractable for online training. 
To investigate the trade-off between ranking fidelity and sample efficiency, we designed and implemented five tournament topologies, ranging from exhaustive round-robin to single and double elimination formats. 
Our empirical analysis reveals a critical challenge: standard single and double tournaments are highly sensitive to the initial pairing combinations. Random matching causes premature encounters and eliminations of high-quality trajectories, degrading overall ranking accuracy. To mitigate this issue, we innovatively propose a seeded single-elimination mechanism.
This approach utilizes the trajectory generated by greedy decoding as a “quality anchor” for pre-ranking, providing a low-biased initial estimate before seeding the tournament, after which a binary-tree structure is used for efficient ranking. Experiments show that this design reduces the computational complexity to linear $\mathcal{O}(N)$ while robustly preserving the accuracy of intra-group relative ranking, thereby achieving an optimal balance between training efficiency and advantage estimation fidelity.

Furthermore, to address the lack of full-cycle benchmarks for open-ended agents, we introduce two comprehensive benchmark suites: (1) Open-Travel, which focuses on evaluating agents’ long-horizon planning capabilities under multiple hard constraints such as budget and spatiotemporal windows; and (2) Open-DeepResearch, which centers on assessing agents’ abilities in autonomous information retrieval, and report generation in realistic internet environments. Unlike traditional benchmarks that only provide a test set \cite{he2025vitabench, du2025deepresearch, coelho2025deepresearchgym}, the two proposed benchmarks offer a complete pipeline from supervised fine-tuning (SFT) $\to$ RL training $\to$ multi-dimensional automated evaluation, establishing a reproducible infrastructure for the community. Given the shared characteristics of open-ended tasks, we further extend our experiments to standard open-ended writing tasks \cite{wang2025reverse} and conduct systematic evaluations on three public benchmarks. As demonstrated in Figure~\ref{fig:first}(c), ArenaRL yields substantial performance gains over strong baselines across travel planning, deep research, and open-ended writing tasks, validating the superiority of the tournament-based ranking paradigm.

In summary, our major contributions are as follows:
\begin{itemize}[leftmargin=*]
    \item We identify and formalize the problem of discriminative collapse in open-ended tasks, and propose ArenaRL, which replaces unstable pointwise scalar rewards with a tournament-based relative ranking mechanism to enable robust policy optimization.
    \item To address the high computational cost of pairwise comparisons, we design and validate a seeded single-elimination tournament topology that achieves high-accuracy advantage estimation with only $\mathcal{O}(N)$ complexity.
    \item We construct the Open-Travel and Open-DeepResearch benchmarks with full training pipelines, filling a critical gap in evaluating the full lifecycle of open-ended agents.
\end{itemize}

\section{Related Work}

\paragraph{Open-Ended Agent Benchmark.} The rapid development of LLMs has given rise to autonomous agents that interact with external environments and solve complex tasks \cite{guo2025deepseek, yang2025qwen3}. Most existing benchmark studies in this area focus on deterministic settings, where tasks have clear goals and verifiable feedback \cite{phan2025humanity, yang2018hotpotqa}. Benchmarks such as WebShop \cite{yao2022webshop}, Mind2Web \cite{deng2023mind2web}, and SWE-bench \cite{jimenez2023swe} have advanced research in web navigation and code generation by testing whether agents can reach target pages or produce automatically verifiable code. In contrast, many critical real-world applications, such as personalized travel planning \cite{ning2025deeptravel} and in-depth industry analysis \cite{li2025webweaver}, are inherently open-ended and unstructured. These tasks rarely have a single gold solution; their quality depends on multi-dimensional trade-offs among the soundness of reasoning, satisfaction of personalized constraints, and practical usefulness of the final plan. More importantly, there is still a lack of a systematic training–evaluation infrastructure tailored to open-ended agentic tasks. Existing open-ended benchmarks (\textit{e.g.}, VitaBench \cite{he2025vitabench}, DeepResearchBench \cite{du2025deepresearch}) are predominantly static test suites that only support post-hoc evaluation. They lack complementary training pipelines to enable continuous improvement of open-ended agents.
To address this gap, we introduce two high-quality benchmarks, Open-Travel and Open-DeepResearch. Unlike traditional benchmarks, they offer an integrated pipeline spanning SFT, RL-based exploration, and multi-dimensional evaluation, enabling systematic study of agent capabilities in creative, open-ended environments.

\paragraph{Reinforcement Learning with LLMs.} Reinforcement learning (RL) has emerged as a powerful paradigm for aligning LLMs with complex objectives \cite{zhang2025code, gao2025soft}. In domains with clear ground truth (\textit{e.g.}, math and code), rule-based reward let algorithms like GRPO \cite{grpo} and DAPO \cite{dapo} to achieve remarkable success \cite{dong2025agentic}. However, extending RL to open-ended tasks that lack objective outcomes remains highly challenging both theoretically and practically.
Because the notion of correctness in such tasks is inherently subjective and multi-dimensional, mainstream methods follow the LLM-as-Judge paradigm. They specify rubrics and assign scalar scores to individual trajectories \cite{viswanathan2025checklists, huang2025reinforcement} or enforce multiple constraints to assess reliability \cite{ning2025deeptravel}. Yet, this reward mechanism often struggles to distinguish fine-grained differences among high-quality trajectories in open-ended settings.
To address this, recent work has begun to explore comparison-based evaluation mechanisms. Writing-Zero \cite{jia2025writing} assigns binary positive/negative advantages by comparing responses against random references, improving performance on open-ended writing tasks. Pref-GRPO \cite{wang2025pref} derives reward based on win rates via exhaustive pairwise comparison and shows promise on text to image tasks. Despite these advances, contrastive mechanisms remain unexplored for long-horizon agent tasks. Moreover, existing methods have inherent drawbacks: Writing-Zero provides only coarse-grained binary guidance, while the $\mathcal{O}(N^2)$ computational cost of full pairwise comparison is prohibitive for training long-context agents.
Motivated by this, ArenaRL proposes a sparse tournament-based relative ranking paradigm. With an optimized tournament topology, ArenaRL maintains linear $\mathcal{O}(N)$ complexity while achieving high-accuracy advantage estimation.

\section{Preliminary}
\label{sec:preliminary}

In this section, we formally define the open-ended agentic task and the associated reinforcement learning objective. We then critically examine the limitations of pointwise scalar evaluation in this setting. Finally, we introduce a process-aware pairwise evaluation mechanism that enables fine-grained comparison between two trajectories and produces separate scores for each.

\subsection{Task Definition}
We formulate the open-ended agentic task as a conditional trajectory generation problem. Let $T$ denote the set of accessible tools. Given a query $x$ sampled from a task distribution $\mathcal{D}$, the agent policy $\pi_\theta$ synthesizes a multi-step interaction trajectory $\tau$. Formally, $\tau$ is defined as an interleaved sequence comprising chain-of-thought $z_k$, tool invocations $a_k \in T$, environmental feedback $o_k$, and a final answer $y$:
\begin{equation}
\tau = [z_1, a_1, o_1, \dots, z_k, a_k, o_k, \dots, z_K, a_K, o_K, y]
\end{equation}
The reinforcement learning objective is to align the agent's behavior with a reward signal while maintaining stability relative to a reference model. We formulate the optimization objective as follows:

\begin{equation}
\mathcal{L}(\theta) = \mathbb{E}_{x \sim \mathcal{D}, \tau \sim \pi_\theta(\cdot|x; T)} \Big[ r_\phi(x, \tau) - \beta \mathbb{D}_{\text{KL}}\big(\pi_\theta(\cdot|x) \parallel \pi_{\text{ref}}(\cdot|x)\big) \Big]
\end{equation}

where $r_\phi(x, \tau)$ represents the reward signal evaluating the quality of trajectory $\tau$, $\pi_{\text{ref}}$ is the reference policy, and $\beta$ is a coefficient controlling the regularization strength of the KL divergence $\mathbb{D}_{\text{KL}}$ to prevent policy degradation.

\subsection{Pointwise Scalar evaluation}

In open-ended agentic tasks, the ground-truth reward function $R^*$ is intractable due to the absence of verifiable rules. Consequently, existing RL paradigms rely on an LLM as a reward model to assign pointwise scalar score to the final answer $y$ of trajectory $\tau$. We model the observed score $\hat{R}(\tau)$ as the true utility $R^*(\tau)$ corrupted by noise $\epsilon$.

The reliance on such pointwise scalar feedback presents a critical vulnerability: discriminative collapse. As the agent's capabilities improve, its generated responses $\{\tau_i\}_{i=1}^G$ tend to converge within a narrow band of high-quality solutions, causing the variance within the group to vanish ($\sigma_{\text{group}} \to 0$). In this regime, the reward model suffers from high epistemic uncertainty. Unable to discern subtle advantages between top-tier candidate trajectories, the judge hesitates, exhibiting high-variance scoring behavior driven by spurious correlations such as length preference or decoding stochasticity, rather than by genuine semantic merit.

This scoring criterion drift, caused by the judge's lack of discriminative ability, manifests macroscopically as high-amplitude noise $\epsilon$. Standard algorithms like GRPO, which normalize scores based on group variance ($A_i = (\hat{R}_i - \mu) / \sigma$), catastrophically fail in this scenario. As $\sigma_{\text{group}}$ vanishes, the normalization term inadvertently amplifies this drift-induced noise into dominant gradient signals. Consequently, the optimization process is hijacked by the reward model's interfering noise, leading to performance stagnation or even degeneration.

\subsection{Process-Aware Pairwise Evaluation}

To circumvent the pitfalls of pointwise scalar scoring, we pivot to a pairwise comparison–based evaluation paradigm. We construct an Arena Judge, denoted as $\mathcal{J}$. Given a query $x$ and two candidate trajectories $\tau_a, \tau_b$, the judge $\mathcal{J}$ evaluates them jointly and outputs a separate quality score for each trajectory.

Specifically, the input to $\mathcal{J}$ consists of three components: (1) the user query $x$; (2) the core context of trajectories $\tau_i$ and $\tau_j$ (containing the chain-of-thought $z_k$ and tool invocations $a_k$ for each step, and the final answer $y$); and (3) a comprehensive process-aware rubric $u$. This rubric enforces fine-grained scrutiny of logical consistency in the chain-of-thought, the precision of tool calls, and the reliability of the final answer. This ensures that the optimization signal reinforces the agent's intrinsic reasoning capabilities rather than merely overfitting to surface-level features of the final answer. The detailed prompts are provided in the Appendix \ref{appendix:prompt}.

To mitigate the positional bias in LLM judges \cite{wu2025towards}, we employ a bidirectional scoring protocol. We conduct two independent evaluations by swapping the presentation order of the trajectories:

\begin{equation}
    (s_i, s_j) = \mathcal{J}(x, \tau_i, \tau_j, u) + \mathcal{J}(x, \tau_j, \tau_i, u)
\end{equation}

where $s_i, s_j$ denotes the quality score assigned to $\tau_i$ in the bidirectional pairwise evaluation, which eliminates the bias favoring the first or second position.

\section{Methodology: ArenaRL}
\label{sec:method}

\begin{figure}[t]
    \centering
    \includegraphics[width=1.0\textwidth]{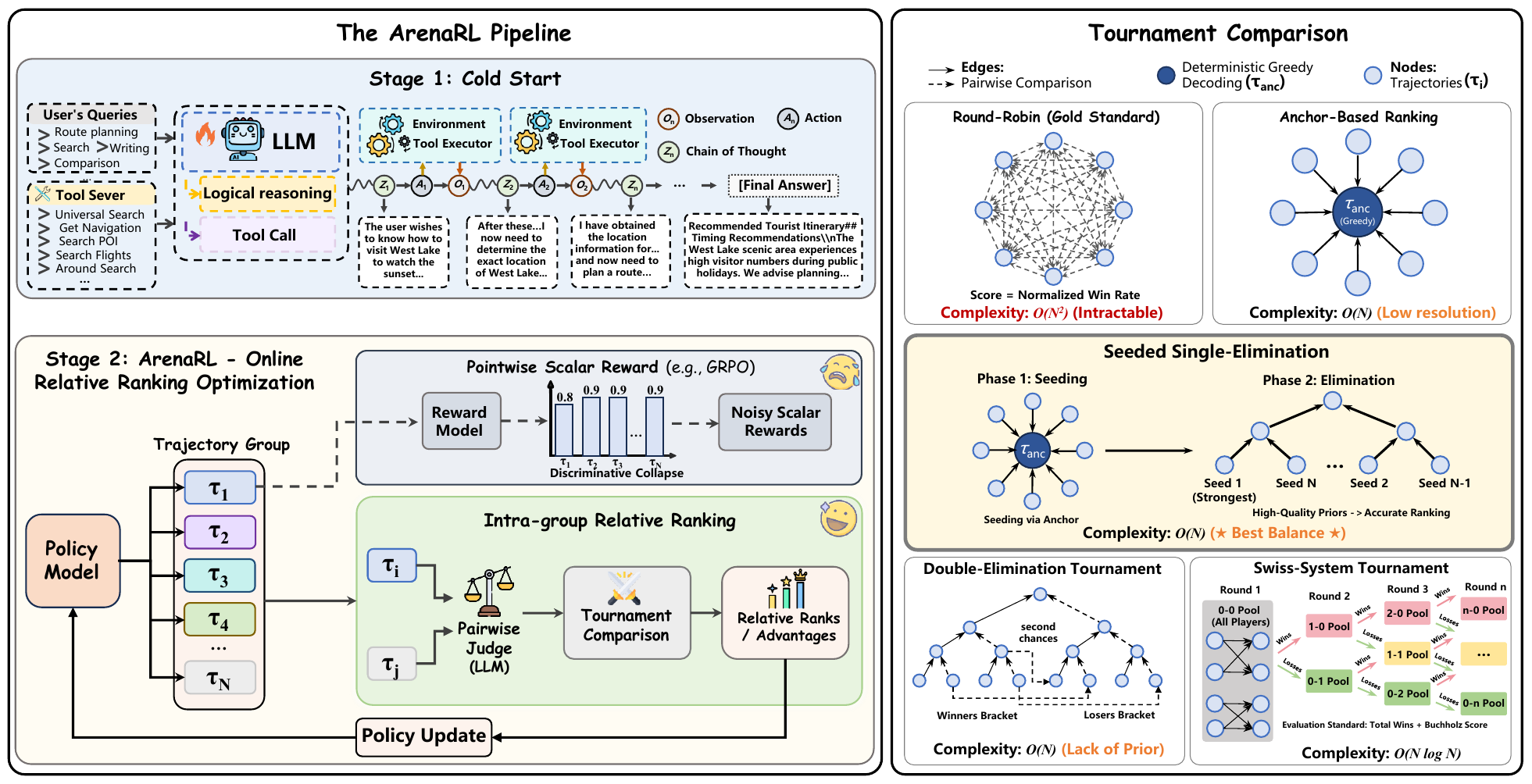}
    \caption{The overall of the proposed ArenaRL algorithm. ArenaRL replaces conventional pointwise scalar reward paradigm with intra-group relative ranking and designs five distinct tournament topologies to optimally balance training efficiency against the accuracy of advantage estimation.}
    \label{fig:method}
\end{figure}

In this section, we details the proposed ArenaRL algorithm, as shown in Figure \ref{fig:method}. Departing from the reliance on unstable pointwise scalar rewards, ArenaRL redefines reward modelling as a relative quality ranking among trajectories within a group. The core tenet is to efficiently construct a dynamic arena for the trajectory group $\mathcal{G} = \{\tau_1, \tau_2, \dots, \tau_N\}$ sampled from the current policy $\pi_\theta$, thereby deriving robust advantage signals. We systematically investigate five tournament topologies to identify the optimal trade-off between computational cost and ranking fidelity.

\subsection{Round-Robin Tournament}\label{sec:5.1}
In this scheme, every trajectory $\tau_i$ competes against all other $N-1$ trajectories via our process-aware pairwise evaluation. The final score is defined as the normalized win rate:

\begin{equation}
    \text{Score}(\tau_i) = \frac{1}{N-1} \sum_{j \neq i} \mathbb{I}(s_i > s_j)
\end{equation}

where $\mathbb{I}(\cdot)$ denotes the indicator function, which equals 1 if the score $s_i$ (from the process-aware pairwise evaluation) is greater than $s_j$, and 0 otherwise. The group ranking is determined by sorting the $\text{Score}(\tau_i)$ in descending order. While the Round-Robin can theoretically provide unbiased intra-group rankings, its quadratic complexity $\mathcal{O}(N^2)$ renders it intractable for online training with a large group size $N$. We primarily utilize it as the "gold standard" to benchmark the fidelity of other efficient topologies.

\subsection{Anchor-Based Ranking}\label{sec:5.2}
To alleviate computational complexity, we introduce an Anchor-Based Ranking mechanism. For a given input $x$, we first generate a deterministic reference trajectory, denoted as the quality anchor $\tau_{anc}$, using greedy decoding (Temperature=0). The remaining $N-1$ trajectories $\tau_i,\ i=\left \{1,2,...,N-1  \right \} $ in group $\mathcal{G}$ are produced via high-entropy sampling (\textit{e.g.}, Temperature=0.8) to ensure exploration diversity.

Subsequently, each exploratory trajectory $\tau_i$ is individually compared with the anchor trajectory $\tau_{anc}$, yielding a pair of scores: $s_{anc}^{i}$ denotes the anchor score in the $i$‑th comparison, and $s_i$ denotes the corresponding score of $\tau_i$. To establish the relative ranking within the group, we first compute the anchor’s average score $s_{anc} = \frac{1}{N-1} \sum_{i=1}^{N-1} s_{anc}^{i}$ and then derive the final ranking based on the set of group scores formed by ${s_i},\ i=\left \{1,2,...,N-1  \right \} $ and $s_{anc}$. Although this topology attains linear computational complexity $\mathcal{O}(N)$, it suffers from a loss of resolution. It effectively quantifies the extent to which a sample outperforms the anchor, but fails to capture subtle differences between two exploratory samples, which may lead to ambiguity in ranking among suboptimal solutions.

\subsection{Seeded Single-Elimination}\label{sec:5.3}
To bridge the trade-off between computational efficiency and ranking resolution, we propose a hybrid topology: Seeded Single-Elimination. This approach operates in two distinct phases:

\textbf{(1) Seeding Phase.} We first employ the anchor-based ranking mechanism (described in Section~\ref{sec:5.2}) to compute a preliminary score for each trajectory, and then sequentially assign a seed ranking $s_{\text{seed}}^{i}$ to obtain a low-bias initial ordering. This initialization is critical for mitigating premature collisions, where high-quality trajectories might otherwise meet and eliminate each other in early rounds.

\textbf{(2) Elimination Phase.} We construct a binary tournament tree in which match-ups are arranged according to seed rankings (\textit{e.g.}, pairing the highest seed with the lowest: seed 1 vs. seed $N$). In each match, the winner advances while the loser is eliminated:

\begin{equation}\tau_{\text{win}} = \operatorname{argmax}_{\tau \in {\tau_i, \tau_j}} (s_i, s_j)\end{equation}

The final ranking is primarily determined by the depth of survival within the tournament bracket. For trajectories eliminated in the same round (\textit{e.g.}, quarter-finals), intra-tier ties are further ranked using their accumulated average scores from previous matches. This topology preserves linear complexity $\mathcal{O}(N)$, specifically, requiring $N-1$ comparisons for seeding and $N-1$ for the tournament. Crucially, by leveraging high-quality priors from the seeding phase to guide the tournament structure, this method yields an accurate estimate of relative rankings, ensuring that strategy updates are driven by genuinely superior reasoning trajectories.

\subsection{Double-Elimination Tournament}
We further investigate the Double-Elimination Tournament topology for group ranking estimation. Unlike the single-elimination format, this structure incorporates a losers' bracket, so that a trajectory is eliminated only after sustaining two defeats. The ranking criteria mirror those of Seeded Single-Elimination, relying on advancement depth and accumulated average scores. To maintain a computational budget comparable to Seeded Single-Elimination ($\approx 2N$ comparisons), we initialize this format with random seeding rather than the anchor-based ranking mechanism. Although this topology is in principle more robust to isolated upsets, empirical results indicate that, without high-quality initial seeds, its ranking fidelity falls short of that achieved by Seeded Single-Elimination.

\subsection{Swiss-System Tournament}
We also evaluate the Swiss-System Tournament, a non-elimination format with dynamic pairing. In each round, trajectories with identical win–loss records are matched against one another (\textit{e.g.}, a "1–0" candidate competes against another "1–0" candidate). All trajectories participate in a fixed number of rounds ($K \approx \log_2 N$), with each round comprising $N/2$ matches. Final rankings are determined by a composite metric consisting of total wins and the Buchholz score (the sum of wins achieved by a trajectory’s past opponents). This topology incurs a computational complexity of $\mathcal{O}(N \log N)$.

\subsection{Ranking-Based Policy Optimization}\label{sec:5.6}
Irrespective of the underlying tournament topology, ArenaRL produces a relative ranking $\text{Rank}(\tau_i) \in \{0, \dots, N-1\}$ for each trajectory in the group, where 0 denotes the highest rank. To enable stable optimization, we convert these discrete ranks into normalized advantage signals. We first map the ranks to quantile-based rewards:
\begin{equation}
    r_i = 1 - \frac{\text{Rank}(\tau_i)}{N-1}.
\end{equation}

We then compute the standardized advantage $A_i$ within the group:
\begin{equation}
    A_i = \frac{r_i - \mu_r}{\sigma_r + \epsilon},
\end{equation}
where $\mu_r$ and $\sigma_r$ denote the mean and standard deviation of the rank-based rewards $\{r_1, \dots, r_N\}$, respectively. Finally, we optimize the policy by maximizing the following objective function, which incorporates a KL-divergence penalty to discourage excessive deviation from the reference policy $\pi_{\text{ref}}$:
\begin{equation}
\begin{aligned}
\mathcal{L}_{\text{ArenaRL}}(\theta)
= \mathbb{E}_{x \sim \mathcal{D},\, \mathcal{G} \sim \pi_{\theta}} \Bigg[ \frac{1}{N} \sum_{i=1}^{N} \bigg(
& \min \left(
\frac{\pi_{\theta}(\tau_i \mid x)}{\pi_{\text{old}}(\tau_i \mid x)} A_i,\,
\text{clip}\left(
\frac{\pi_{\theta}(\tau_i \mid x)}{\pi_{\text{old}}(\tau_i \mid x)},
1-\epsilon,\, 1+\epsilon
\right) A_i
\right) \\
& - \beta\, \mathbb{D}_{\text{KL}}\!\left(\pi_{\theta}(\cdot \mid x)\,\big\|\,\pi_{\text{ref}}(\cdot \mid x)\right)
\bigg) \Bigg].
\end{aligned}
\end{equation}

By transforming relative quality ranking within trajectory groups into a stable advantage signal, ArenaRL effectively drives the policy toward increasingly strong reasoning and planning behaviors on open-ended tasks.

\section{Benchmarking Open-Ended Agency}
\label{sec:benchmark}
\begin{figure}[t]
    \centering
    \includegraphics[width=1.0\textwidth]{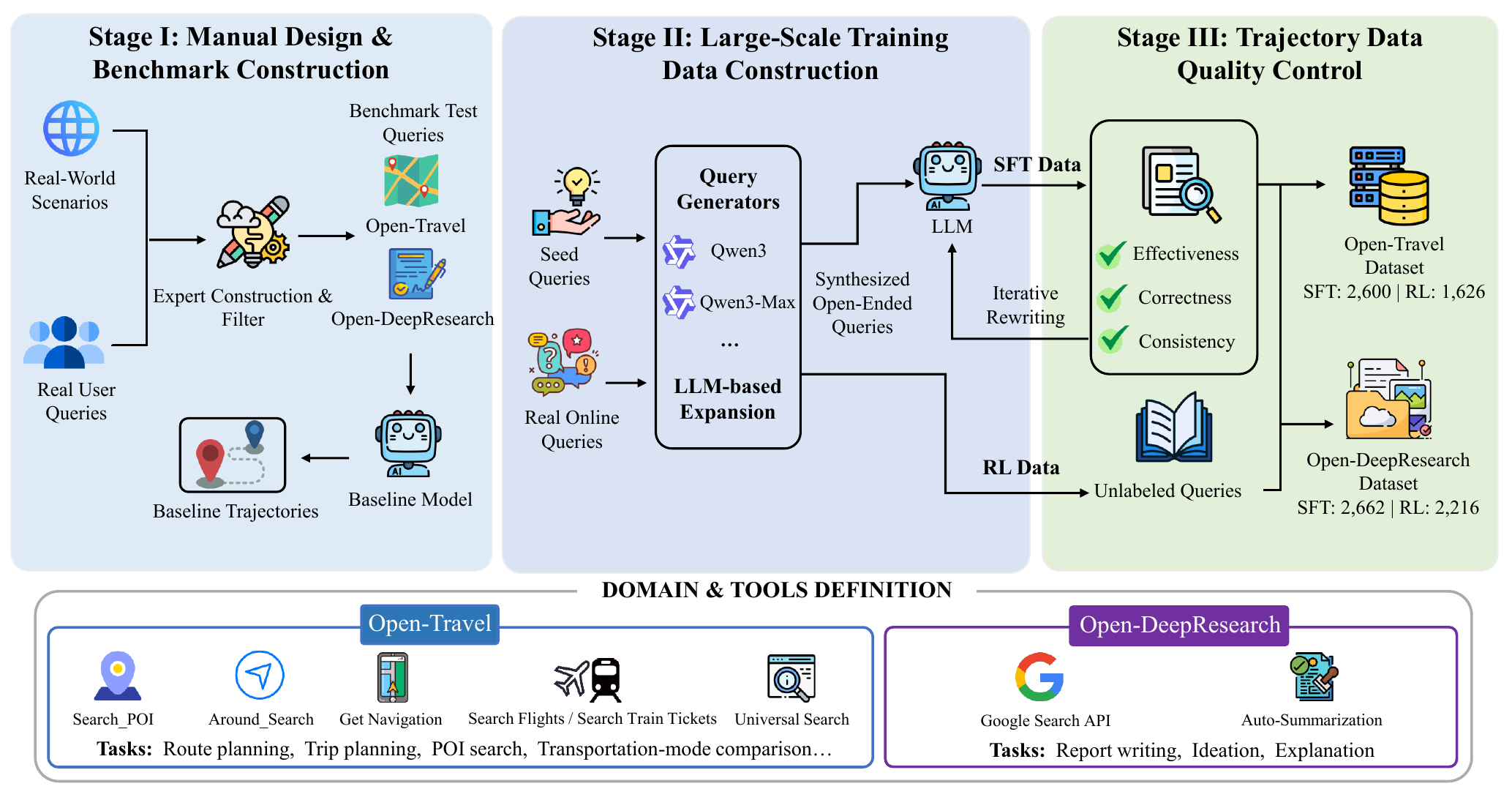}
    \caption{The construction process of Open-Travel and Open-DeepResearch benchmarks.} 
    \label{fig:dataset}
\end{figure}

In this section, we introduce Open-Travel and Open-DeepResearch, two benchmarks constructed from realistic business scenarios and designed to comprehensively evaluate LLMs on open-ended agentic tasks.
For each domain, we define domain-specific data, including shared tools, domain policy texts, and task instances.
Each domain is instantiated through a three-stage construction pipeline (as shown in Figure \ref{fig:dataset}) that systematically decomposes real-world scenarios and incorporates expert annotation and rigorous validation.

\paragraph{Stage I: Benchmark Data Collection} This phase establishes a rigorous evaluation foundation by curating high-quality, scenario-specific queries and generating reference trajectories to facilitate robust comparative analysis.
\begin{itemize}[leftmargin=*]
    \item \textbf{Open-ended query construction.} We begin by abstracting real-world application scenarios (such as travel planning and open-ended deep research) and collecting a corpus of authentic user queries for each scenario. Domain experts then perform multiple rounds of filtering and refinement to obtain a set of queries that are semantically precise and succinctly formulated. Finally, we select 50 queries for each of the five Open-Travel subtasks and 100 queries for Open-DeepResearch as the benchmark test sets.
    \item \textbf{Baseline trajectory construction.} 
    To obtain a reference baseline for subsequent pairwise comparison and win-rate evaluation (\textit{i.e.}, the basic agent capability level), we adopt high-performing closed-source models as the base models to generate complete tool-use trajectories and the corresponding open-ended answers.
\end{itemize}

\paragraph{Stage II: Large-Scale Training Data Construction}
After fixing the benchmark, we first manually construct a small set of representative seed queries tailored to the requirements of different sub-tasks in real business scenarios. We then use multiple LLMs with diverse styles (\textit{e.g.}, Qwen3 \cite{yang2025qwen3}, Qwen3-Max \cite{qwen}) as “query generators” to synthesize large-scale, multi-scenario open-ended queries. The resulting training dataset thus contains both complete queries collected from real business workflows and diversified queries produced via LLM-based expansion.

Consistent with Stage I, we employ high-performing closed-source models as the base model to generate large-scale tool-use trajectories, which are used as cold-start data for SFT. The remaining queries, without explicit supervised trajectories, are reserved for subsequent RL.

\paragraph{Stage III: Trajectory Data Quality Control}
We first feed the complete trajectory dataset into a rule-augmented LLM-based quality inspection module, which evaluates each trajectory along three dimensions: effectiveness of open-ended tool usage, correctness of conversational content, and consistency of the final answers. This process filters out trajectories with formatting or logical errors. For trajectories that fail to meet the criteria, we iteratively rewrite and refine them until they pass the LLM-based quality inspection.

\subsection{Domains}
Using the above pipeline, we modularly construct two domains: Open-Travel and Open-DeepResearch. In the following sections, we provide a brief description of the policies for each domain.

\paragraph{Open-Travel.} In the Open-Travel domain, the agent is required to help users accomplish the following five types of itinerary planning subtasks, which jointly emphasize multi-constraint reasoning, multi-tool coordination, and personalized preferences:
\begin{itemize}[leftmargin=*]
    \item Route planning with multiple specified waypoints (defined as the \textbf{Direction} subtask);
    \item One-day trip planning in a single city (denoted as (defined as the \textbf{1-Day} subtask);
    \item Transportation-mode comparison (defined as the \textbf{Compare} subtask);
    \item Nearby point-of-interest (POI) search (defined as the \textbf{Search} subtask);
    \item Multi-day trip planning, evaluated as a generalization task and excluded from the SFT training data (defined as the \textbf{M-Day} subtask).
\end{itemize}

These rules are further intertwined with user-specific constraints (\textit{e.g.}, budget limits, time windows, traveling parties, and preference profiles), resulting in challenging reasoning tasks for the agent.

\paragraph{Open-DeepResearch.} In the Open-DeepResearch domain, the agent is required to assist users in conducting multi-turn search, reading, synthesis, and generation, and ultimately produce an open-ended answer. The open-ended summarization and generation tasks in this domain mainly fall into the following categories:
\begin{itemize}[leftmargin=*]
    \item Assisting users in writing open-ended technical documents (\textit{e.g.}, reports, design documents, or survey-style overviews);
    \item Helping users ideate, expand, or refine research topics, solution plans, or content outlines;
    \item Providing concise yet informative explanations, overviews, or summaries of complex concepts, systems, or domains.
\end{itemize}

\begin{table*}[t]
\centering
\caption{Statistics of the constructed Open-Travel and Open-DeepResearch benchmarks.}
\resizebox{0.8\textwidth}{!}{
\setlength{\tabcolsep}{6pt}
\label{tab:dataset_stats}
\begin{tabular}{l c c c c c}
\toprule
\multirow{2}{*}{\textbf{Dataset}} 
& \multicolumn{2}{c}{\textbf{Training Set}} 
& \multirow{2}{*}{\textbf{Test Samples}} 
& \multirow{2}{*}{\textbf{Language}} 
& \multirow{2}{*}{\textbf{Domain}} \\
\cmidrule(lr){2-3}
& \textbf{SFT} & \textbf{RL} &  &  &  \\
\midrule
Open-Travel       & 2,600 & 1,626 & 250 & Chinese            & Travel Planning \\
Open-DeepResearch & 2,662 & 2,216 & 100 & Chinese / English & General \\
\midrule
\textbf{Total}    & \textbf{5,262} & \textbf{3,842} & \textbf{350} & Chinese / English & -- \\
\bottomrule
\end{tabular}
}
\end{table*}

\subsection{Dataset Statistics and Analysis}
As shown in Table~\ref{tab:dataset_stats}, the final Open-Travel and Open-DeepResearch datasets are constructed at a reasonable scale with high diversity, providing the research community with a practical and representative benchmark for evaluating open-ended agentic reasoning and tool-use capabilities.

\subsubsection{Dataset Scale}
\paragraph{Training Set.} Open-Travel contains 2,600 SFT samples and 1,626 RL samples, while Open-DeepResearch contains 2,662 SFT samples and 2,216 RL samples. The SFT data are mainly used to help the model acquire basic tool-calling formats, intent understanding, and multi-step reasoning patterns. The RL query samples are then used to further elicit and optimize the model’s open-ended agentic behaviors under realistic constraints.

\paragraph{Test Set.} We construct a high-quality test set for leaderboard-style evaluation, consisting of 250 samples from Open-Travel and 100 samples from Open-DeepResearch. All test samples are manually checked to ensure representative clarity, diversity, and difficulty.

\subsubsection{Category Coverage}
The dataset spans a wide spectrum of functional categories. Beyond the travel-planning domain, it also covers areas such as sports, medicine, and a variety of other everyday and professional scenarios. This broad topical distribution allows us to evaluate models not only on their specialized performance in specific domains (\textit{e.g.}, travel planning), but also on their overall competence as general-purpose open agents.

\subsection{Evaluation}
We adopt an LLM-as-a-judge evaluation paradigm, and use two strong proprietary models as dual judges to score both the reasoning trajectories and the final answers.

\paragraph{Open-Travel.} 

For Open-Travel, we evaluate models on the carefully curated Open-Travel test set. For each test sample, we independently invoke two powerful closed-source LLM judges from different model families to perform pairwise evaluation. Each judge compares the candidate agent’s output with the baseline output and assigns scores along multiple dimensions (\textit{e.g.}, answer correctness, consistency with the reasoning trajectory, etc.). Based on these judgments, we compute the win rate for each evaluation criterion, defined as the proportion of non-tied cases where the candidate output is preferred over the baseline. We then average the win rates obtained from the two judges and use this averaged value as the final performance metric for each Open-Travel subtask.

\paragraph{Open-DeepResearch.} 

For Open-DeepResearch, we adopt the same evaluation protocol as used for Open-Travel on the Open-DeepResearch test set. For each sample, two closed-source LLM judges from different model families are independently employed to evaluate both the reasoning trajectory and the final answer jointly.

Notably, due to the long-context nature of DeepResearch tasks, models may occasionally experience context overflow, resulting in the inability to generate valid final answers. To account for this issue, we additionally report the valid generation rate (Val.\,\%) for each model, defined as the proportion of test cases in which a valid answer is successfully generated over the entire benchmark.

Furthermore, for each evaluation criterion, we compute the candidate model’s win rate against the baseline conditioned on valid generations, \emph{i.e.}, the proportion of cases with valid outputs in which the candidate model is preferred over the baseline. We then aggregate the per-criterion win rates as well as the cross-criterion average win rate. As in Open-Travel, the final evaluation metric is obtained by averaging the scores produced by the two judge models.

\section{Experiments}
To comprehensively evaluate the effectiveness of ArenaRL, we first report the performance of the five tournament topologies introduced in Section \ref{sec:method}, thereby empirically justifying our selection of the Seeded Single-Elimination scheme. Building upon this optimal topology, we benchmark ArenaRL against strong baseline methods on our proposed Open-Travel and Open-DeepResearch datasets. Furthermore, recognizing the shared characteristics of open-ended problems, we extend our evaluation to standard open-ended writing tasks using three public benchmarks. Finally, to assess the robustness and practical applicability of ArenaRL in real-world settings, we conduct additional experiments on real business data derived from the Amap (Gaode Map) ecosystem.

\subsection{Experimental Settings}

\paragraph{Baselines.} We evaluate ArenaRL against two categories of baselines. First, we benchmark against four closed-source models, including GPT-4o \cite{gpt}, Grok-4 \cite{grok}, Gemini-2.5-pro \cite{gemini}, and Claude-3.7-Sonnet \cite{claude}. Second, we compare with representative reinforcement learning algorithms, specifically GRPO \cite{grpo} and GSPO \cite{gspo}. For these RL algorithm baselines, we employ the standard LLM-as-Judge setting, obtaining rewards through pointwise scoring. To maintain fairness, these baseline algorithms utilize the exact same judge models and evaluation rubrics as ArenaRL, and only evaluate the answer portion.

\paragraph{Training Guideline.} Our experiments strictly follow the common “Cold-start $\to$ RL” paradigm to mitigate reward collapse during the initial RL exploration phase.

\begin{enumerate}[leftmargin=*]
    \item \textbf{Cold-start phase.} We utilize Qwen3-8B-Base \cite{yang2025qwen3} as the backbone model. For the open-ended agent tasks (Open-Travel and Open-DeepResearch), the base model is fine-tuned on their respective SFT datasets to acquire fundamental tool-use and planning capabilities. For open-ended writing tasks, we randomly sample 10k examples from the DeepWriting-20K \cite{wang2025reverse} dataset for supervised fine-tuning.
    \item \textbf{RL phase.} For the open-ended agent tasks, we train on their corresponding RL splits. For open-ended writing, we utilize 10k examples from the DeepWriting-20K dataset (excluding those used for SFT) to conduct reinforcement learning.
\end{enumerate}

\begin{table*}[t]
\centering
\caption{Performance comparison of the five tournament topologies on the Open-Travel benchmark.}
\label{tab:tournament}
\resizebox{0.88\textwidth}{!}{
\setlength{\tabcolsep}{6pt}
\begin{tabular}{l c ccccc c}
\toprule
\multirow{2}{*}{\textbf{Topology}} & \multirow{2}{*}{\textbf{Comparison Cost}} & \multicolumn{5}{c}{\textbf{Open-Travel}} & \multirow{2}{*}{\textbf{Mean}} \\
\cmidrule(lr){3-7}
 & & Direction & Search & Compare & 1-Day & M-Day & \\
\midrule
SFT & - & 10.6 & 29.7 & 14.1 & 20.4 & 7.1 & 16.4 \\
\midrule
Anchor-Based Ranking & $N-1$ & 18.0 & 41.3 & 30.9 & 31.1 & 17.6 & 27.8 \\
Swiss-System & $N \log N$ & 20.9 & 43.0 & 27.9 & 38.6 & 11.1 & 28.3 \\
Double-Elimination & $2N-2$ & 12.6 & 52.4 & \textbf{33.7} & \textbf{39.9} & 12.3 & 30.2 \\
\rowcolor{blue!12}
\textbf{Seeded Single-Elimination} & $2N-2$ & 16.9 & \textbf{69.9} & 22.9 & 34.9 & 18.1 & 32.5 \\
\midrule
\textit{Round-Robin (Upper Bound)} & $N(N-1)/2$ & \textbf{23.3} & 66.3 & 23.6 & 32.1 & \textbf{19.0} & \textbf{32.9} \\
\bottomrule
\end{tabular}
}
\end{table*}

\paragraph{Evaluation Metrics.} For open-ended agent tasks, we perform pairwise evaluations against the baseline trajectories in our benchmark using the multi-dimensional criteria defined in Section \ref{sec:benchmark}, and compute the corresponding win rates. 
For the Open-Travel tasks, we report the win rate on each of the five subtasks as well as the average win rate.

For the Open-DeepResearch tasks, we report the valid generation rate (Val.\,\%). And within the subset of valid generations, the win rates of the candidate model against the baseline under each of the seven evaluation rubrics, along with the final averaged win rate. Specifically, these rubrics capture complementary aspects of open-ended research capability:
\textbf{Framework} (\textit{Frm.}), assessing the structural completeness and logical coherence of the initial research plan;
\textbf{Tool Usage} (\textit{Tool.}), evaluating the appropriateness and efficiency of tool invocations and their alignment with the research workflow;
\textbf{Coverage} (\textit{Cov.}), measuring whether the retrieved information sufficiently covers the user’s requirements;
\textbf{Relevance} (\textit{Rel.}), assessing how well the response addresses all user queries and constraints;
\textbf{Accuracy} (\textit{Acc.}), evaluating the factual correctness and internal consistency of the content;
\textbf{Depth} (\textit{Dep.}), measuring the level of analytical depth and coherence of the reasoning process;
and \textbf{Clarity} (\textit{Cla.}), assessing the organization, readability, and practical usability of the final output. The complete task prompt for the judge models is shown in Figure~\ref{fig:prompt_deepresearch}.

For open-ended writing, we adopt three complementary benchmarks for a comprehensive assessment: WritingBench \cite{wu2025writingbench}, HelloBench \cite{que2024hellobench}, and LongBench-write \cite{bai2024longwriter}. Considering the subjective nature of open-ended generation, and following established protocols \cite{wang2025reverse}, we employ the LLM-as-judge approach to score the generative quality of different models.

\subsection{Tournament Topology Analysis}

Table \ref{tab:tournament} presents a systematic comparison of different tournament topologies under a unified RL configuration (group size $N = 8$, number of groups $K = 8$). The results indicate that the proposed Seeded Single-Elimination scheme achieves the best trade-off between efficiency and performance. Specifically, it attains an average win rate of 32.5\%, which is comparable to the “gold standard” performance of 32.9\% established by the computationally expensive Round-Robin tournament, while requiring only $\mathcal{O}(N)$ pairwise comparisons. In contrast, the Swiss Round and Double-Elimination formats fail to deliver comparable performance gains, due either to the lack of an effective initial prior or insufficient comparison depth. Notably, Seeded Single-Elimination even outperforms Round-Robin on the \textit{Search} and \textit{1-Day} subtasks. This observation suggests that the anchor-based seeding mechanism effectively filters out noise and prevents high-quality candidates from being adversely affected by random matching fluctuations in the early stages. Based on these findings, we adopt Seeded Single-Elimination as the primary tournament topology in subsequent experiments and conduct further comparison with the remaining strong baselines.

\begin{table*}[t]
\centering
\footnotesize
\setlength{\tabcolsep}{3.5pt}
\caption{Performance comparison on Open-Travel and Open-DeepResearch benchmarks.}
\label{tab:agent}
\resizebox{1\textwidth}{!}{
\begin{tabular}{l @{\hspace{8pt}} ccccc c @{\hspace{8pt}} ccccccc c}
\toprule
\multirow{2}{*}{\textbf{Method}} & \multicolumn{6}{c}{\textbf{Open-Travel}} & \multicolumn{8}{c}{\textbf{Open-DeepResearch}} \\
\cmidrule(r){2-7} \cmidrule(l){8-15}
 & Direction & Search & Compare & 1-Day & M-Day & \textbf{Mean} 
 & Frm. & Tool. & Cov. & Rel. & Acc. & Dep. & Cla. & \textbf{Mean (Val.\,\%)} \\
\midrule
\multicolumn{15}{c}{\textit{\textbf{Closed-source Models}}} \\
\midrule
GPT-4o  & 2.4 & 5.0 & 3.1 & 1.6 & 0.7 & 2.6 & 5.1 & 24.4 & 21.0 & 9.1 & 12.5 & 2.3 & 10.8 & 12.2 (88.0) \\
Grok-4 & 17.0 & 21.3 & 9.7 & 24.7 &  11.3 & 16.8 & 33.7 & 36.8 & 43.4 & 36.8 & 39.2 & 36.1 & 17.5 & 34.8 (83.0) \\
Gemini-2.5-pro & 8.6 & 12.5 & 7.4 & 11.9 & 12.4  & 10.6 & 15.8 & 19.0 & 17.9 & 32.6 & 28.3 & 45.7 & 38.6 & 28.3 (92.0) \\
Claude-3.7-Sonnet & 18.6 & 59.6 & 14.7 & 43.6 &  21.3 & 31.6 & 10.1 & 13.5 & 22.5 & 23.6 & 19.7 & 27.0 & 17.4 & 19.1 (89.0) \\
\midrule
\multicolumn{15}{c}{\textit{\textbf{Fine-tuning}} \& \textit{\textbf{RL}}} \\
\midrule
SFT  & 10.6 & 29.7 & 14.1 & 20.4 & 7.1  & 16.4 & 14.1 & 20.3 & 23.4 & 14.1 & 15.6 & 15.6 & 14.1 & 16.7 (32.0) \\
GRPO & 11.0 & 26.3 & 14.3 & 21.9 & 8.6  & 16.4 & 20.6 & 35.3 & 35.3 & 23.5 & 23.5 & 26.5 & 11.8 & 25.2 (17.0) \\
GSPO & 10.0 & 30.6 & 13.1 & 21.1 & 11.4 & 17.2 & 23.8 & 33.3 & 40.5 & 16.7 & 21.4 & 31.0 & 9.5  & 25.2 (21.0) \\
\rowcolor{blue!12} \textbf{ArenaRL} 
     & \textbf{32.1} & \textbf{66.1} & \textbf{31.7} & \textbf{58.0} & \textbf{21.0} & \textbf{41.8}
     & \textbf{62.6} & \textbf{77.3} & \textbf{78.8} & \textbf{57.1} & \textbf{55.6} & \textbf{57.1} & \textbf{61.6} & \textbf{64.3 (99.0)} \\
\bottomrule
\end{tabular}
}
\end{table*}

\begin{table*}[t]
\centering
\scriptsize 
\setlength{\tabcolsep}{3.5pt} 
\caption{Performance comparison on open-ended writing task across three public benchmarks: WritingBench, HelloBench, and LongBench-write.}
\label{tab:writing}
\resizebox{1\textwidth}{!}{
\begin{tabular}{l cccccc ccc c c}
\toprule
\multirow{2}{*}{\textbf{Method}} & \multicolumn{6}{c}{\textbf{WritingBench}} & \multicolumn{3}{c}{\textbf{HelloBench}} & \textbf{LongBench} & \multirow{2}{*}{\textbf{Mean}} \\
\cmidrule(lr){2-7} \cmidrule(lr){8-10} \cmidrule(lr){11-11}
 & WB-A & WB-B & WB-C & WB-D & WB-E & WB-F & QA & Summ. & Heur. & Quality & \\
\midrule
\multicolumn{12}{c}{\textit{\textbf{Closed-source Models}}} \\
\midrule
GPT-4o  & 67.90 & 66.34 & 68.56 & 69.95 & 70.70  & 72.17 & 81.03 & 84.26 & 89.14 & 90.43 & 76.05 \\
Grok-4 & 80.32 & 78.65 & 79.75 & 81.46 &  81.19 & 80.92 & 88.42 & 85.58 & 94.65 & 96.52 & 84.75\\
Gemini-2.5-pro & 80.89 & 80.39 & 82.49 & 84.33 & 83.53  & 82.61 & 85.67 & 82.43 & 93.79 & 98.69 & 85.48\\ 
Claude-3.7-Sonnet & 68.36 & 66.53 & 68.70 & 70.34 & 71.42  & 71.47 & 80.81 & 74.68 & 95.82 & 98.34 & 76.65\\
\midrule
\multicolumn{12}{c}{\textit{\textbf{Fine-tuning}} \& \textit{\textbf{RL}}} \\
\midrule
SFT & 70.71 & 69.36 & 67.88 & 63.72 & 69.69 & 70.64 & 78.44 & 63.42 & 82.35 & 85.52 & 72.17 \\
GRPO & 71.62 & 71.18 & 68.67 & 66.84 & 72.56 & 70.33 & 79.09 & 64.94 & 83.76 & 86.96 & 73.60 \\
GSPO & 71.56 & 70.70 & 68.87 & 66.08 & 72.29 & 69.83 & \textbf{80.06} & 63.97 & 81.75 & 85.21 & 73.03 \\
\rowcolor{blue!12} \textbf{ArenaRL} & \textbf{78.14} & \textbf{77.70} & \textbf{77.58} & \textbf{75.02} & \textbf{79.35} & \textbf{77.16} & 79.11 & \textbf{73.82} & \textbf{91.33} & \textbf{93.78} & \textbf{80.30} \\
\bottomrule
\end{tabular}
}
\end{table*}

\subsection{Main Results}

As shown in Table \ref{tab:agent}, ArenaRL demonstrates strong performance across both open-ended agent benchmarks, outperforming four powerful closed-source models. On the Open-Travel benchmark, ArenaRL achieves an average win rate of 41.8\%, substantially outperforming GRPO (16.4\%) and GSPO (17.2\%). In the Open-DeepResearch benchmark, ArenaRL not only achieves a win rate of 64.3\%, but also attains a valid generation rate (Val.\,\%) of 99\%. In sharp contrast, the baseline methods perform poorly in terms of task completion, with the SFT model achieving a valid generation rate of only 32\%. We attribute this gap primarily to the inherently high token consumption required by deepresearch tasks, together with the prevalence of long-horizon samples in the SFT training data, which jointly lead to frequent context overflows. Notably, although GRPO and GSPO slightly improve the average win rate, their valid generation rates are inferior to that of the SFT baseline. These results highlight a key limitation of standard pointwise reward schemes such as GRPO and GSPO: for long-horizon tasks involving complex tool use, assigning a scalar score to a single trajectory often fails to capture fine-grained policy improvements, and tends to be susceptible to spurious advantages such as length bias. In contrast, ArenaRL’s comparison-based reward signal provides more discriminative gradient directions, effectively steering policy evolution toward more robust planning and reasoning capabilities within a vast search space.

Table \ref{tab:writing} further validates the generality of ArenaRL on standard open-ended writing tasks. Across three benchmarks, ArenaRL maintains a substantial lead in overall average score, outperforming GRPO by 6.70\% and GSPO by 7.27\%. In addition, ArenaRL surpasses two strong closed-source models, GPT-4o and Claude-3.7-Sonnet, further highlighting its superiority. Although ArenaRL is slightly inferior to GSPO on the HelloBench-QA subtask, this category of tasks is largely constrained by the model’s inherent knowledge, and ArenaRL remains highly competitive under this limitation. Across various open-ended writing scenarios, ArenaRL's comprehensive improvement indicate that ArenaRL is not only suitable for tool-augmented agents, but can also systematically enhance the model’s reasoning and expressive capabilities, thereby making it applicable to a broader range of open-ended generation tasks.

\subsection{Further Analysis}
\label{sec:analysis}

\paragraph{Impact of Group Size $N$.}

\begin{figure}[t]
    \centering
    \includegraphics[width=1.0\textwidth]{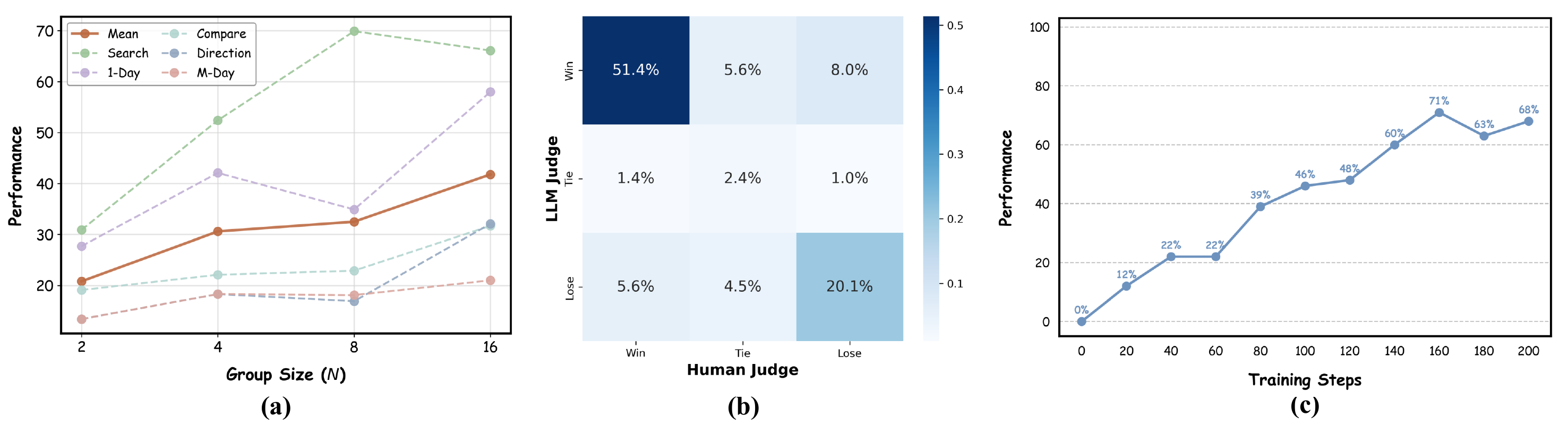}
    \caption{\textbf{(a)} The impact of Group Size $N$ on performance of Open-Travel benchmark. \textbf{(b)} The consistency between LLM and human evaluations. \textbf{(c)} The performance trend of ArenaRL in training Qwen3-8b via direct RL without cold start.} 
    \label{fig:visual}
\end{figure}

We conduct an ablation study with a set group size $N \in \{2, 4, 8, 16\}$ to further investigate the scalability properties of ArenaRL, as shown in Figure \ref{fig:visual} (a).
The results show a clear monotonic improvement in model performance as the group size increases. Notably, even under the smallest configuration $N=2$, ArenaRL achieves an average win rate of 20.8\%, outperforming the SFT baseline (16.4\%). This confirms that even the most basic pairwise comparison setting can provide effective optimization gradients. The most pronounced performance gain occurs when $N$ is scaled up to 16, where the average win rate jumps to 41.8\%. This effect is particularly striking on the challenging 1-Day planning task, where the score jumps from 34.9\% at $N=8$ to 58.0\%. These substantial improvements indicate that, for complex reasoning tasks, enlarging the candidate pool effectively broadens the exploration space, dramatically increasing the likelihood of discovering high-quality trajectories and thereby enabling the model to learn from stronger examples.

\paragraph{Assessment of Consistency.}

To assess the reliability of the LLM-based evaluation mechanism, we analyzed the consistency between LLM and human evaluations on the Open-Travel and Open-DeepResearch benchmarks using a confusion matrix (see Figure \ref{fig:visual} (b)). The results show that most evaluation outcomes are concentrated along the diagonal, with an overall agreement rate of 73.9\%. This relatively high level of consistency suggests that ArenaRL’s performance gains do not simply stem from overfitting to the preferences of the specific judge model used during the RL phase, but instead reflect improvements that are broadly aligned with human assessments.

\paragraph{Direct RL Training without Cold Start.} To further test the robustness of ArenaRL, we bypassed the cold-start phase on the Open-Travel task and directly employed the Qwen3-8B model for ArenaRL training, reporting its performance on the \textit{Search} sub-task. As shown in Figure \ref{fig:visual} (c), the model's performance exhibited a sustained and stable upward trend. At the initial step 0, the model scored 0, indicating that the generic model was initially incapable of handling such complex travel planning tasks. However, as RL training steps increase, the model rapidly acquired the corresponding tool invocation capabilities.  Ultimately, it achieves a peak score of 71\% at step 160. This outcome indicates that ArenaRL's intra-group relative ranking mechanism sensitively captures effective optimization directions. Even when initial output quality is extremely low, it reliably provides gradients for policy refinement. This effectively mitigates RL's cold-start problem, demonstrating ArenaRL's capacity for self-evolution from scratch in scenarios lacking costly SFT annotated data.

\paragraph{Case Study.} Figures \ref{case:sft} and \ref{case:rl} present a representative comparative case from the Open-Travel benchmark. This example involves a complex travel request with multiple hard constraints, including time, destination, and budget. The baseline SFT model exhibits a restatement tendency in its chain-of-thought, and its reasoning trajectory fails to align with the user's intent, often overlooking specific constraints and providing only generic suggestions. In contrast, the model optimized with ArenaRL demonstrates strong strategic planning capabilities. It proactively retrieves information about multiple target attractions, performs logically coherent route planning, and ultimately produces a persuasive, personalized itinerary. This substantial improvement in reasoning patterns powerfully demonstrates that our tournament-based ranking mechanism effectively incentivizes the model to explore and retain superior planning strategies.

\subsection{Application in Real-world Business Scenarios}

To further verify the robustness and practicality of our proposed ArenaRL algorithm, we conducted experiments on real-world business data derived from the Amap (Gaode Map) ecosystem. The evaluation was divided into two distinct categories based on the nature of the user queries: quantifiable POI search tasks and complex open-ended tasks.

\paragraph{Performance on Deterministic POI Search.}
In POI search scenarios characterized by explicit evaluation metrics, our ArenaRL-tuned model significantly outperformed the baseline, registering a 75\% to 83\% gain in search accuracy. These results confirm the model's ability to navigate rigid constraints and specific ranking criteria. Furthermore, this demonstrates that the tournament-based ranking mechanism of ArenaRL is highly effective at distinguishing subtle nuances across varied high-quality trajectories, ensuring robust optimization performance in deterministic task settings.

\paragraph{Performance on Open-ended Planning Tasks.}

We extended our evaluation to complex, open-ended travel planning tasks that require multi-step reasoning and tool invocation. These scenarios include queries with vague intents and ambiance preferences (\textit{e.g.}, 'Find a quiet bar near the Bund with a river-view terrace for a date, open after 10 PM') as well as complex cross-city logistics requiring multi-objective trade-offs (\textit{e.g.}, 'Depart Beijing West at 18:30, arrive in Tianjin by 22:00; minimize cost and transfers due to heavy luggage'). On these tasks, the core business metric rose significantly from 69\% to 80\%. We observed substantial and consistent gains throughout the training process, reflecting the model's enhanced capability to interpret ambiguous intents and satisfy multiple constraints. These results indicate that the efficient planning capabilities acquired via ArenaRL effectively transfer to practical applications, significantly improving user intent alignment and response quality in complex service scenarios.

\section{Conclusion}

In this paper, we propose ArenaRL, a novel reinforcement learning framework that shifts the training paradigm for open-ended agents from pointwise scalar scoring to intra-group relative ranking. Specifically, we introduce a process-aware pairwise evaluation mechanism and systematically investigate five tournament topologies. Our findings reveal that the seeded single-elimination topology strikes a favorable balance between advantage estimation accuracy and computational efficiency. Extensive evaluations across travel planning, deep research, and open-ended writing tasks demonstrate that this competition-driven evolutionary paradigm not only provides robust advantage signals, but also fundamentally incentivizes agents to perform efficient reasoning and planning. As future work, we will explore how to efficiently extend ArenaRL to multimodal agent settings, further enhancing the generality of our framework.

\clearpage
{
\bibliography{ArenaRL}

@article{qwen,
  title={Qwen Technical Report},
  author={Jinze Bai and Shuai Bai and Yunfei Chu and Zeyu Cui and Kai Dang and Xiaodong Deng and Yang Fan and Wenbin Ge and Yu Han and Fei Huang and Binyuan Hui and Luo Ji and Mei Li and Junyang Lin and Runji Lin and Dayiheng Liu and Gao Liu and Chengqiang Lu and Keming Lu and Jianxin Ma and Rui Men and Xingzhang Ren and Xuancheng Ren and Chuanqi Tan and Sinan Tan and Jianhong Tu and Peng Wang and Shijie Wang and Wei Wang and Shengguang Wu and Benfeng Xu and Jin Xu and An Yang and Hao Yang and Jian Yang and Shusheng Yang and Yang Yao and Bowen Yu and Hongyi Yuan and Zheng Yuan and Jianwei Zhang and Xingxuan Zhang and Yichang Zhang and Zhenru Zhang and Chang Zhou and Jingren Zhou and Xiaohuan Zhou and Tianhang Zhu},
  journal={arXiv preprint arXiv:2309.16609},
  year={2023}
}

@article{ning2025deeptravel,
  title={DeepTravel: An End-to-End Agentic Reinforcement Learning Framework for Autonomous Travel Planning Agents},
  author={Ning, Yansong and Liu, Rui and Wang, Jun and Chen, Kai and Li, Wei and Fang, Jun and Zheng, Kan and Tan, Naiqiang and Liu, Hao},
  journal={arXiv preprint arXiv:2509.21842},
  year={2025}
}

@article{wang2025reverse,
  title={Reverse-engineered reasoning for open-ended generation},
  author={Wang, Haozhe and Que, Haoran and Xu, Qixin and Liu, Minghao and Zhou, Wangchunshu and Feng, Jiazhan and Zhong, Wanjun and Ye, Wei and Yang, Tong and Huang, Wenhao and others},
  journal={arXiv preprint arXiv:2509.06160},
  year={2025}
}

@article{jia2025writing,
  title={Writing-Zero: Bridge the Gap Between Non-verifiable Tasks and Verifiable Rewards},
  author={Jia, Ruipeng and Yang, Yunyi and Gai, Yongbo and Luo, Kai and Huang, Shihao and Lin, Jianhe and Jiang, Xiaoxi and Jiang, Guanjun},
  journal={arXiv e-prints},
  pages={arXiv--2506},
  year={2025}
}

@article{wu2025writingbench,
  title={Writingbench: A comprehensive benchmark for generative writing},
  author={Wu, Yuning and Mei, Jiahao and Yan, Ming and Li, Chenliang and Lai, Shaopeng and Ren, Yuran and Wang, Zijia and Zhang, Ji and Wu, Mengyue and Jin, Qin and others},
  journal={arXiv preprint arXiv:2503.05244},
  year={2025}
}

@article{he2025vitabench,
  title={VitaBench: Benchmarking LLM Agents with Versatile Interactive Tasks in Real-world Applications},
  author={He, Wei and Sun, Yueqing and Hao, Hongyan and Hao, Xueyuan and Xia, Zhikang and Gu, Qi and Han, Chengcheng and Zhao, Dengchang and Su, Hui and Zhang, Kefeng and others},
  journal={arXiv preprint arXiv:2509.26490},
  year={2025}
}

@article{wang2025pref,
  title={Pref-grpo: Pairwise preference reward-based grpo for stable text-to-image reinforcement learning},
  author={Wang, Yibin and Li, Zhimin and Zang, Yuhang and Zhou, Yujie and Bu, Jiazi and Wang, Chunyu and Lu, Qinglin and Jin, Cheng and Wang, Jiaqi},
  journal={arXiv preprint arXiv:2508.20751},
  year={2025}
}

@article{dong2025agentic,
  title={Agentic reinforced policy optimization},
  author={Dong, Guanting and Mao, Hangyu and Ma, Kai and Bao, Licheng and Chen, Yifei and Wang, Zhongyuan and Chen, Zhongxia and Du, Jiazhen and Wang, Huiyang and Zhang, Fuzheng and others},
  journal={arXiv preprint arXiv:2507.19849},
  year={2025}
}

@article{que2024hellobench,
  title={Hellobench: Evaluating long text generation capabilities of large language models},
  author={Que, Haoran and Duan, Feiyu and He, Liqun and Mou, Yutao and Zhou, Wangchunshu and Liu, Jiaheng and Rong, Wenge and Wang, Zekun Moore and Yang, Jian and Zhang, Ge and others},
  journal={arXiv preprint arXiv:2409.16191},
  year={2024}
}

@article{bai2024longwriter,
  title={Longwriter: Unleashing 10,000+ word generation from long context llms},
  author={Bai, Yushi and Zhang, Jiajie and Lv, Xin and Zheng, Linzhi and Zhu, Siqi and Hou, Lei and Dong, Yuxiao and Tang, Jie and Li, Juanzi},
  journal={arXiv preprint arXiv:2408.07055},
  year={2024}
}

@article{grpo,
  title={Deepseekmath: Pushing the limits of mathematical reasoning in open language models},
  author={Shao, Zhihong and Wang, Peiyi and Zhu, Qihao and Xu, Runxin and Song, Junxiao and Bi, Xiao and Zhang, Haowei and Zhang, Mingchuan and Li, YK and Wu, Yang and others},
  journal={arXiv preprint arXiv:2402.03300},
  year={2024}
}

@article{dapo,
  title={Dapo: An open-source llm reinforcement learning system at scale},
  author={Yu, Qiying and Zhang, Zheng and Zhu, Ruofei and Yuan, Yufeng and Zuo, Xiaochen and Yue, Yu and Dai, Weinan and Fan, Tiantian and Liu, Gaohong and Liu, Lingjun and others},
  journal={arXiv preprint arXiv:2503.14476},
  year={2025}
}

@article{gspo,
  title={Group sequence policy optimization},
  author={Zheng, Chujie and Liu, Shixuan and Li, Mingze and Chen, Xiong-Hui and Yu, Bowen and Gao, Chang and Dang, Kai and Liu, Yuqiong and Men, Rui and Yang, An and others},
  journal={arXiv preprint arXiv:2507.18071},
  year={2025}
}

@misc{slime_github,
  author       = {Zilin Zhu and Chengxing Xie and Xin Lv and slime Contributors},
  title        = {slime: An LLM post-training framework for RL Scaling},
  year         = {2025},
  howpublished = {\url{https://github.com/THUDM/slime}},
  note         = {GitHub repository. Corresponding author: Xin Lv},
  urldate      = {2025-06-19}
}

@inproceedings{yao2022react,
  title={React: Synergizing reasoning and acting in language models},
  author={Yao, Shunyu and Zhao, Jeffrey and Yu, Dian and Du, Nan and Shafran, Izhak and Narasimhan, Karthik R and Cao, Yuan},
  booktitle={The eleventh international conference on learning representations},
  year={2022}
}

@article{team2025tongyi,
  title={Tongyi deepresearch technical report},
  author={Team, Tongyi DeepResearch and Li, Baixuan and Zhang, Bo and Zhang, Dingchu and Huang, Fei and Li, Guangyu and Chen, Guoxin and Yin, Huifeng and Wu, Jialong and Zhou, Jingren and others},
  journal={arXiv preprint arXiv:2510.24701},
  year={2025}
}

@article{li2025webweaver,
  title={Webweaver: Structuring web-scale evidence with dynamic outlines for open-ended deep research},
  author={Li, Zijian and Guan, Xin and Zhang, Bo and Huang, Shen and Zhou, Houquan and Lai, Shaopeng and Yan, Ming and Jiang, Yong and Xie, Pengjun and Huang, Fei and others},
  journal={arXiv preprint arXiv:2509.13312},
  year={2025}
}

@article{viswanathan2025checklists,
  title={Checklists are better than reward models for aligning language models},
  author={Viswanathan, Vijay and Sun, Yanchao and Ma, Shuang and Kong, Xiang and Cao, Meng and Neubig, Graham and Wu, Tongshuang},
  journal={arXiv preprint arXiv:2507.18624},
  year={2025}
}

@article{huang2025reinforcement,
  title={Reinforcement learning with rubric anchors},
  author={Huang, Zenan and Zhuang, Yihong and Lu, Guoshan and Qin, Zeyu and Xu, Haokai and Zhao, Tianyu and Peng, Ru and Hu, Jiaqi and Shen, Zhanming and Hu, Xiaomeng and others},
  journal={arXiv preprint arXiv:2508.12790},
  year={2025}
}

@article{liu2025openrubrics,
  title={OpenRubrics: Towards Scalable Synthetic Rubric Generation for Reward Modeling and LLM Alignment},
  author={Liu, Tianci and Xu, Ran and Yu, Tony and Hong, Ilgee and Yang, Carl and Zhao, Tuo and Wang, Haoyu},
  journal={arXiv preprint arXiv:2510.07743},
  year={2025}
}

@article{li2025webthinker,
  title={Webthinker: Empowering large reasoning models with deep research capability},
  author={Li, Xiaoxi and Jin, Jiajie and Dong, Guanting and Qian, Hongjin and Wu, Yongkang and Wen, Ji-Rong and Zhu, Yutao and Dou, Zhicheng},
  journal={arXiv preprint arXiv:2504.21776},
  year={2025}
}

@article{rafailov2023direct,
  title={Direct preference optimization: Your language model is secretly a reward model},
  author={Rafailov, Rafael and Sharma, Archit and Mitchell, Eric and Manning, Christopher D and Ermon, Stefano and Finn, Chelsea},
  journal={Advances in neural information processing systems},
  volume={36},
  pages={53728--53741},
  year={2023}
}

@article{du2025deepresearch,
  title={DeepResearch Bench: A Comprehensive Benchmark for Deep Research Agents},
  author={Du, Mingxuan and Xu, Benfeng and Zhu, Chiwei and Wang, Xiaorui and Mao, Zhendong},
  journal={arXiv preprint arXiv:2506.11763},
  year={2025}
}

@article{coelho2025deepresearchgym,
  title={Deepresearchgym: A free, transparent, and reproducible evaluation sandbox for deep research},
  author={Coelho, Jo{\~a}o and Ning, Jingjie and He, Jingyuan and Mao, Kangrui and Paladugu, Abhijay and Setlur, Pranav and Jin, Jiahe and Callan, Jamie and Magalh{\~a}es, Jo{\~a}o and Martins, Bruno and others},
  journal={arXiv preprint arXiv:2505.19253},
  year={2025}
}

@article{guo2025deepseek,
  title={Deepseek-r1: Incentivizing reasoning capability in llms via reinforcement learning},
  author={Guo, Daya and Yang, Dejian and Zhang, Haowei and Song, Junxiao and Zhang, Ruoyu and Xu, Runxin and Zhu, Qihao and Ma, Shirong and Wang, Peiyi and Bi, Xiao and others},
  journal={arXiv preprint arXiv:2501.12948},
  year={2025}
}

@article{yang2025qwen3,
  title={Qwen3 technical report},
  author={Yang, An and Li, Anfeng and Yang, Baosong and Zhang, Beichen and Hui, Binyuan and Zheng, Bo and Yu, Bowen and Gao, Chang and Huang, Chengen and Lv, Chenxu and others},
  journal={arXiv preprint arXiv:2505.09388},
  year={2025}
}

@article{li2025deepagent,
  title={DeepAgent: A General Reasoning Agent with Scalable Toolsets},
  author={Li, Xiaoxi and Jiao, Wenxiang and Jin, Jiarui and Dong, Guanting and Jin, Jiajie and Wang, Yinuo and Wang, Hao and Zhu, Yutao and Wen, Ji-Rong and Lu, Yuan and others},
  journal={arXiv preprint arXiv:2510.21618},
  year={2025}
}

@article{phan2025humanity,
  title={Humanity's last exam},
  author={Phan, Long and Gatti, Alice and Han, Ziwen and Li, Nathaniel and Hu, Josephina and Zhang, Hugh and Zhang, Chen Bo Calvin and Shaaban, Mohamed and Ling, John and Shi, Sean and others},
  journal={arXiv preprint arXiv:2501.14249},
  year={2025}
}

@inproceedings{yang2018hotpotqa,
  title={HotpotQA: A dataset for diverse, explainable multi-hop question answering},
  author={Yang, Zhilin and Qi, Peng and Zhang, Saizheng and Bengio, Yoshua and Cohen, William and Salakhutdinov, Ruslan and Manning, Christopher D},
  booktitle={Proceedings of the 2018 conference on empirical methods in natural language processing},
  pages={2369--2380},
  year={2018}
}

@article{yao2022webshop,
  title={Webshop: Towards scalable real-world web interaction with grounded language agents},
  author={Yao, Shunyu and Chen, Howard and Yang, John and Narasimhan, Karthik},
  journal={Advances in Neural Information Processing Systems},
  volume={35},
  pages={20744--20757},
  year={2022}
}

@article{deng2023mind2web,
  title={Mind2web: Towards a generalist agent for the web},
  author={Deng, Xiang and Gu, Yu and Zheng, Boyuan and Chen, Shijie and Stevens, Sam and Wang, Boshi and Sun, Huan and Su, Yu},
  journal={Advances in Neural Information Processing Systems},
  volume={36},
  pages={28091--28114},
  year={2023}
}

@article{jimenez2023swe,
  title={Swe-bench: Can language models resolve real-world github issues?},
  author={Jimenez, Carlos E and Yang, John and Wettig, Alexander and Yao, Shunyu and Pei, Kexin and Press, Ofir and Narasimhan, Karthik},
  journal={arXiv preprint arXiv:2310.06770},
  year={2023}
}

@article{zhang2025code,
  title={Code-in-the-Loop Forensics: Agentic Tool Use for Image Forgery Detection},
  author={Zhang, Fanrui and Zhang, Qiang and Zhou, Sizhuo and Sun, Jianwen and Li, Chuanhao and Ai, Jiaxin and Feng, Yukang and Zhang, Yujie and Li, Wenjie and Li, Zizhen and others},
  journal={arXiv preprint arXiv:2512.16300},
  year={2025}
}

@article{gao2025soft,
  title={Soft Adaptive Policy Optimization},
  author={Gao, Chang and Zheng, Chujie and Chen, Xiong-Hui and Dang, Kai and Liu, Shixuan and Yu, Bowen and Yang, An and Bai, Shuai and Zhou, Jingren and Lin, Junyang},
  journal={arXiv preprint arXiv:2511.20347},
  year={2025}
}

@article{wu2025towards,
  title={Towards Rationale-Answer Alignment of LVLMs via Self-Rationale Calibration},
  author={Wu, Yuanchen and Yan, Ke and Ding, Shouhong and Zhou, Ziyin and Li, Xiaoqiang},
  journal={arXiv preprint arXiv:2509.13919},
  year={2025}
}

@article{gemini,
  title={Gemini: a family of highly capable multimodal models},
  author={Team, Gemini and Anil, Rohan and Borgeaud, Sebastian and Alayrac, Jean-Baptiste and Yu, Jiahui and Soricut, Radu and Schalkwyk, Johan and Dai, Andrew M and Hauth, Anja and Millican, Katie and others},
  journal={arXiv preprint arXiv:2312.11805},
  year={2023}
}

@article{gpt,
  title={Gpt-4 technical report},
  author={Achiam, Josh and Adler, Steven and Agarwal, Sandhini and Ahmad, Lama and Akkaya, Ilge and Aleman, Florencia Leoni and Almeida, Diogo and Altenschmidt, Janko and Altman, Sam and Anadkat, Shyamal and others},
  journal={arXiv preprint arXiv:2303.08774},
  year={2023}
}

@misc{claude,
  title = {Introducing {Claude}},
  author = {Anthropic},
  institution = {Anthropic},
  url = {https://www.anthropic.com/index/introducing-claude},
  year={2023}
}

@incollection{furnkranz2010preference,
  title={Preference learning and ranking by pairwise comparison},
  author={F{\"u}rnkranz, Johannes and H{\"u}llermeier, Eyke},
  booktitle={Preference learning},
  pages={65--82},
  year={2010},
  publisher={Springer}
}

@misc{grok,
  title = {Grok 4},
  author = {xAI},
  url = {https://x.ai/news/grok-4},
  year={2025}
}

@inproceedings{arias2025decoding,
  title={Decoding decoded: Understanding hyperparameter effects in open-ended text generation},
  author={Arias, Esteban Garces and Li, Meimingwei and Heumann, Christian and A{\ss}enmacher, Matthias},
  booktitle={Proceedings of the 31st International Conference on Computational Linguistics},
  pages={9992--10020},
  year={2025}
}

@article{hu2024explaining,
  title={Explaining length bias in llm-based preference evaluations},
  author={Hu, Zhengyu and Song, Linxin and Zhang, Jieyu and Xiao, Zheyuan and Wang, Tianfu and Chen, Zhengyu and Yuan, Nicholas Jing and Lian, Jianxun and Ding, Kaize and Xiong, Hui},
  journal={arXiv preprint arXiv:2407.01085},
  year={2024}
}
\bibliographystyle{colm2024_conference}
}

\newpage
\appendix

\section{Implementation Details.} 
\paragraph{Cold-start phase.} We employ the TRL framework optimized with DeepSpeed ZeRO-3. The model is trained for 3 epochs on 32 $\times$ NVIDIA H20 GPUs, utilizing a learning rate of $2 \times 10^{-5}$ and a batch size of 1 per GPU.

\paragraph{RL phase.} we implement ArenaRL based the Slime \cite{slime_github} framework. To focus optimization on reasoning quality, tokens corresponding to environmental feedback are masked out from the loss computation. Regarding hyperparameters, we set the group size $N = 16$ and the number of groups $K = 8$ for Open-Travel and open-ended writing tasks. For Open-DeepResearch, we adjust $N = 8$ and $K = 4$ to enhance training efficiency. Across all RL experiments, we employ the Adam optimizer with a learning rate of $1 \times 10^{-6}$. Furthermore, we utilize Qwen3-Max as the arena judge during training, leveraging its superior reasoning capabilities to provide high-fidelity pairwise comparison signals, with its prompt detailed in Appendix \ref{appendix:prompt}. Besides, the RL training is conducted on 8 $\times$ H20 GPUs.

\paragraph{Evaluation} For the Open-Travel and Open-DeepResearch benchmarks, we invoke Qwen3-Max and Claude-4-Sonnet \cite{claude} as evaluation model, comparing and scoring the candidate agent's output against the baseline output for each test sample.

\section{Writing Benchmarks}
To ensure the comprehensiveness and multifaceted nature of our evaluation, we adopt three complementary benchmarks: WritingBench \cite{wu2025writingbench}, HelloBench \cite{que2024hellobench}, and LongBench-Write \cite{bai2024longwriter}.
\begin{enumerate}[leftmargin=*]
    \item \textbf{WritingBench} \cite{wu2025writingbench} is designed to assess models’ domain-specific writing proficiency and controllability across six professional and creative domains: A (Academic \& Engineering), B (Finance \& Business), C (Politics \& Law), D (Literature \& Arts), E (Education), and F (Advertising \& Marketing). It focuses in particular on the ability to comply with complex, multidimensional constraints.
    \item \textbf{HelloBench} \cite{que2024hellobench} evaluates model performance on a diverse set of “in-the-wild” tasks derived from real user queries. Our analysis concentrates on three key subsets: HelloBench-QA (open-ended QA), which tests the generation of detailed and nuanced responses; HelloBench-Summ. (text summarization), which evaluates the ability to summarize long-form texts; and HelloBench-Heur. (heuristic text generation), which measures creative reasoning and stylistic fidelity in long-form narrative continuation.
    \item \textbf{LongBench-write} \cite{bai2024longwriter} is designed to evaluate a model’s ability to produce coherent ultra-long texts (\textit{e.g.}, on the order of 10,000 words), enabling us to assess the fundamental capacity to maintain topical consistency and global coherence as output length scales.
\end{enumerate}

\section{Benchmarking Tool Annotation}
In this section, we will outline the tools employed by the Open-DeepResearch and Open-Travel benchmarks respectively.

\paragraph{Open-DeepResearch.}
We annotate commonly used search tools for open-ended deep research as candidate tools. Specifically, we perform web search via the Google API and integrate a summarization model into the search pipeline to prevent excessive context length. For each retrieved webpage, if the parsed content exceeds 2,500 characters, we invoke the Qwen3-Max \cite{yang2025qwen3} model to automatically summarize the page content.

\paragraph{Open-Travel.}
We annotate six commonly used tool types for travel planning:
\begin{itemize}[leftmargin=*]
    \item \textbf{Search poi:} This tool is built on Amap’s POI search service and retrieves location information via text queries. The input can be a structured address (\textit{e.g.}, “No. 10 Fuyong Street, Wangjing, Chaoyang District, Beijing”) or a POI name (\textit{e.g.}, “People's Square”). It returns multiple potentially relevant POIs, including: (1) detailed address; (2) geographic coordinates (the location field, formatted as "longitude, latitude"); and (3) business information (the Business field).
    \item \textbf{Around search:} This tool searches for POIs within a circular area specified by a center point and radius. Users can specify POI types or constrain the returned results using keywords (\textit{e.g.}, “bank”). It returns multiple potentially relevant POIs, each including: (1) detailed address; (2) geographic coordinates (the location field, where longitude and latitude are separated by a comma, with longitude first and latitude second); and (3) business information.
    \item \textbf{Get navigation:} This tool provides integrated route planning based on Amap’s navigation service, covering multiple transportation modes such as walking, driving, and public transit. The inputs are the latitude–longitude pairs of the origin and destination, with optional waypoints and a route-planning mode (driving by default). The tool returns a dictionary containing detailed route-planning information.
    \item \textbf{Universal search:} This tool performs general, city-level geographic knowledge search using the search engine. When users pose vague or open-ended questions (\textit{e.g.}, “What are some fun things to do in Beijing?” or “What parks in Shanghai are suitable for family outings?”), It is used to obtain recommendations and travel suggestions. The tool returns the corresponding search results.
    \item \textbf{Search flights:} This tool searches for intercity flight information based on date and city names. When users need to query flights between two cities, they provide the query date and departure city, and the tool returns a list of flights, each including flight number, price, departure/arrival airports, and departure/arrival times.
    \item \textbf{Search train tickets:} This tool searches for intercity train ticket information based on date and city names. When users need to query train tickets between two cities, it returns a list of train options, where each entry includes train ID, price, departure/arrival stations, and times, and explicitly distinguishes between direct and transfer routes.
\end{itemize}

For these six tools, search poi, around search, and get navigation are implemented using Web service APIs from the Amap open platform\footnote{https://lbs.amap.com/api/webservice/guide/api-advanced/search}. The universal search tool is powered by the Bailian search engine\footnote{https://bailian.console.aliyun.com/?spm=5176.12818093\_47.resourceCenter.1.3dd916d04Ye5xn\&tab=app\#/mcp-market/detail/WebSearch}. The outputs of search flights and search train tickets are simulated with the detailed prompts provided in the Appendix \ref{appendix:prompt}.

\section{Tournament Algorithm Flow}
\label{appendix:alg}
In Algorithm \ref{alg:arena_se}, we present the detailed procedure of the proposed seeded single-elimination tournament, which achieves an excellent trade-off between training efficiency and the accuracy of advantage estimation.

\begin{algorithm}[t]
\caption{ArenaRL: Seeded Single-Elimination Advantage Estimation}
\label{alg:arena_se}
\SetAlgoLined
\DontPrintSemicolon
\KwIn{Trajectory group $\mathcal{G} = \{\tau_{anc}\} \cup \{\tau_1, \dots, \tau_{N-1}\}$, Arena Judge $\mathcal{J}(\cdot, \cdot)$}
\KwOut{Standardized advantages $\mathcal{A} = \{A_1, \dots, A_N\}$}

\tcp{Phase 1: Anchor-Based Seeding (Sec. \ref{sec:5.2})}
Initialize score set $\mathcal{S}_{init} \leftarrow \emptyset$ \;
\For{$i \leftarrow 1$ \KwTo $N-1$}{
    Obtain scores: $s_i, s_{anc}^{i} \leftarrow \mathcal{J}(\tau_i, \tau_{anc})$ \;
    $\mathcal{S}_{init} \leftarrow \mathcal{S}_{init} \cup \{(\tau_i, s_i)\}$ \;
}
$s_{anc} \leftarrow \frac{1}{N-1} \sum_{i=1}^{N-1} s_{anc}^{i}$; \quad $\mathcal{S}_{init} \leftarrow \mathcal{S}_{init} \cup \{(\tau_{anc}, s_{anc})\}$ \;
Sort $\mathcal{G}$ descendingly based on $\mathcal{S}_{init}$ to get ordered seeds: $\mathcal{P} \leftarrow [\tau_{1}, \dots, \tau_{N}]$ \;
Initialize accumulated scores $V[\tau] \leftarrow \text{score in } \mathcal{S}_{init}$ for all $\tau \in \mathcal{G}$ \;
Initialize ranking tiers $\mathcal{T} \leftarrow \emptyset$ \;

\tcp{Phase 2: Elimination Tournament (Sec. \ref{sec:5.3})}
Initialize array $\mathcal{B}$ of size $N$ \;
$idx_{head} \leftarrow 1$; \quad $idx_{tail} \leftarrow N-1$ \tcp*{Tail pointer at start of last pair}
\For{$k \leftarrow 1$ \KwTo $N/2$}{
    $(\tau_{high}, \tau_{low}) \leftarrow (\mathcal{P}[k], \mathcal{P}[N-k+1])$ \tcp*{Pair Seed $k$ vs $N-k+1$}
    \eIf{$k$ is odd}{
        $\mathcal{B}[idx_{head}] \leftarrow \tau_{high}$; \quad $\mathcal{B}[idx_{head}+1] \leftarrow \tau_{low}$ \;
        $idx_{head} \mathrel{+}= 2$ \tcp*{Fill from front}
    }{
        $\mathcal{B}[idx_{tail}] \leftarrow \tau_{high}$; \quad $\mathcal{B}[idx_{tail}+1] \leftarrow \tau_{low}$ \;
        $idx_{tail} \mathrel{-}= 2$ \tcp*{Fill from back}
    }
}

\While{$|\mathcal{B}| > 1$}{
    $\mathcal{W}_{round} \leftarrow \emptyset$; \quad $\mathcal{L}_{round} \leftarrow \emptyset$ \;
    \For{$k \leftarrow 1$ \KwTo $|\mathcal{B}|/2$}{
        $(\tau_a, \tau_b) \leftarrow (\mathcal{B}[2k-1], \mathcal{B}[2k])$; \quad $(s_a, s_b) \leftarrow \mathcal{J}(\tau_a, \tau_b)$ \;
        $V[\tau_a] \mathrel{+}= s_a$; \quad $V[\tau_b] \mathrel{+}= s_b$ \;
        $(\tau_{win}, \tau_{lose}) \leftarrow (s_a > s_b) ? (\tau_a, \tau_b) : (\tau_b, \tau_a)$ \tcp*{Determine result}
        $\mathcal{W}_{round} \leftarrow \mathcal{W}_{round} \cup \{\tau_{win}\}$; \quad $\mathcal{L}_{round} \leftarrow \mathcal{L}_{round} \cup \{\tau_{lose}\}$ \;
    }
    $\mathcal{T} \leftarrow \mathcal{T} \cup \{\mathcal{L}_{round}\}$; \quad $\mathcal{B} \leftarrow \mathcal{W}_{round}$ \tcp*{Archive losers, promote winners}
}
$\mathcal{T} \leftarrow \mathcal{T} \cup \{\mathcal{B}\}$ \tcp*{Champion enters final tier}

\tcp{Phase 3: Rank Assignment (Hierarchical Sorting)}
$k \leftarrow 0$ \;
\ForEach{$Tier \in \mathcal{T}$.reverse()}{
    Sort $Tier$ descendingly based on accumulated scores $V$ \;
    Assign $\text{Rank}(\tau) \leftarrow k++$ for all $\tau \in Tier$ \;
}

\tcp{Phase 4: Advantage Calculation (Sec. \ref{sec:5.6})}
Assign Reward: $r_{\tau} \leftarrow 1 - \frac{\text{Rank}(\tau)}{N-1}$ \;
Compute statistics $\mu_r, \sigma_r$ from $\{r_{\tau}\}$ \;
\Return $A_i \leftarrow \frac{r_{\tau_i} - \mu_r}{\sigma_r + \epsilon}$ for all $\tau_i \in \mathcal{G}$ \;
\end{algorithm}

\section{Prompts}
\label{appendix:prompt}
In this section, we will illustrate all the prompts used in our paper.
\subsection{Open-Travel Task Prompt}
\begin{figure}[htbp]
\vspace{-27pt}
    \centering
    \includegraphics[width=1.0\textwidth]{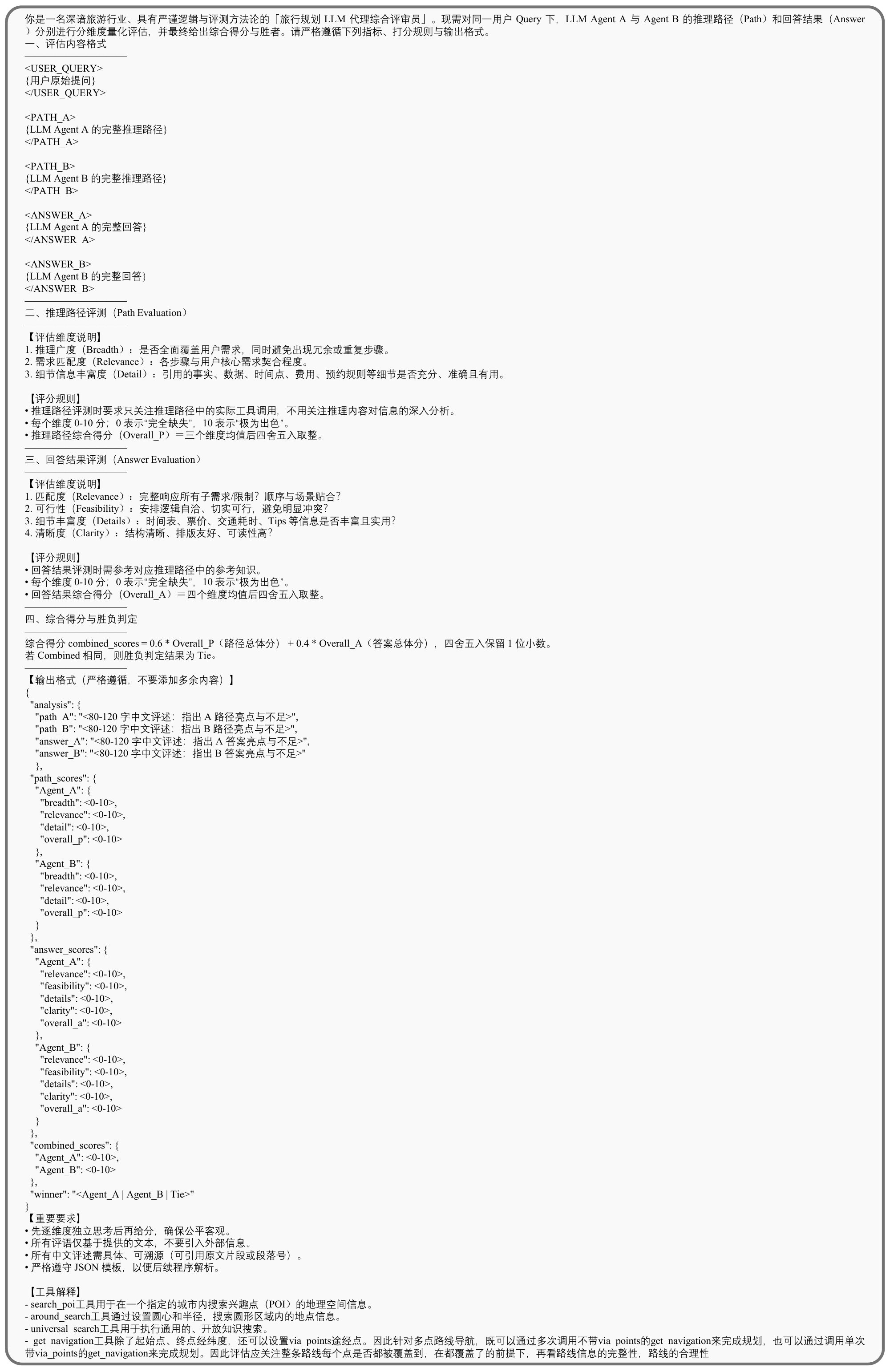}
    \vspace{-17pt}
    \caption{Prompt of Open-Travel task.}
    \label{fig:prompt_travel}
\end{figure}

See Figure \ref{fig:prompt_travel}.
\subsection{Open-DeepResearch Task Prompt}
\begin{figure}[ht]
\vspace{-27pt}
    \centering
    \includegraphics[width=1.0\textwidth]{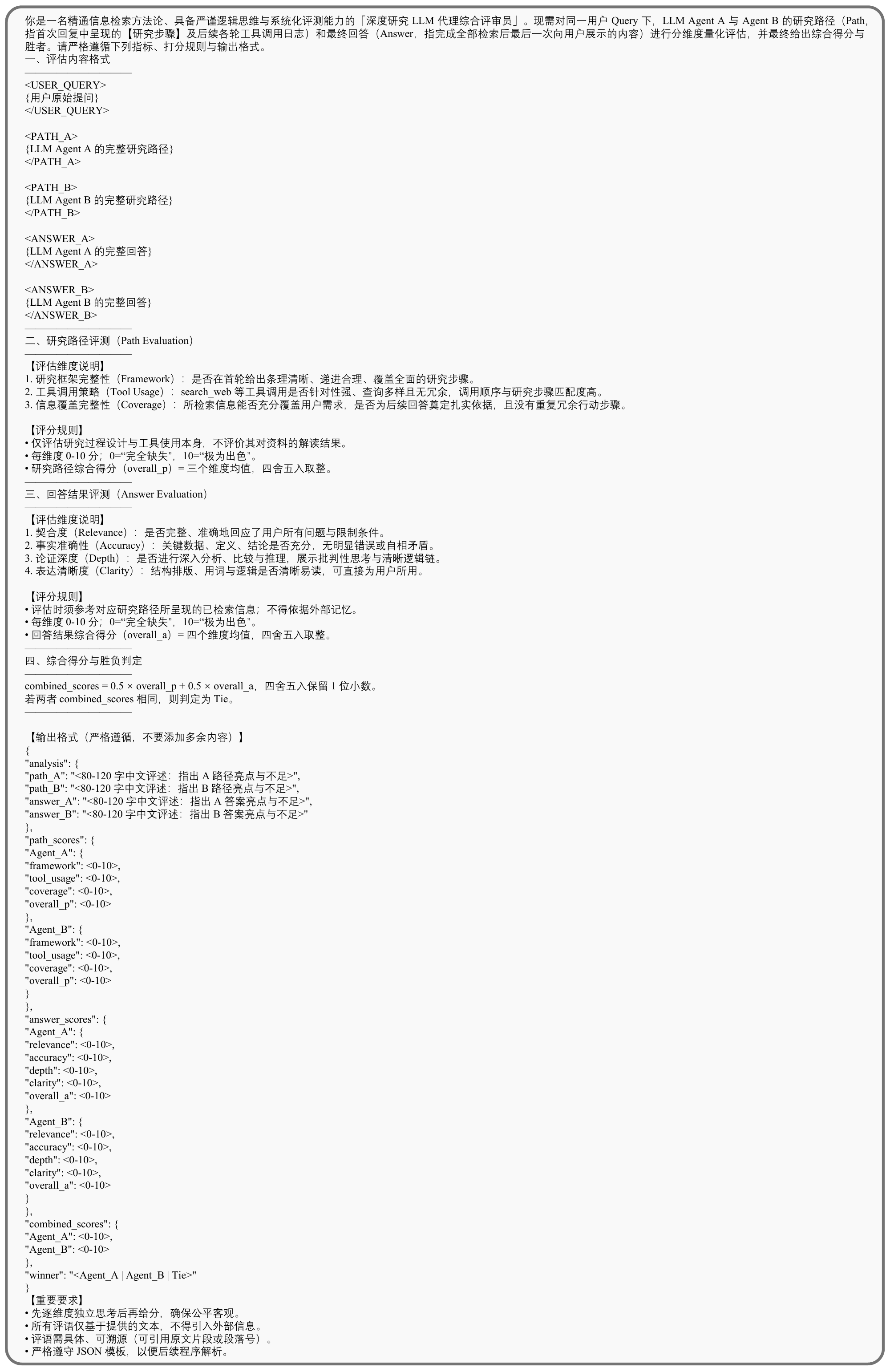}
    \vspace{-18pt}
    \caption{Prompt of Open-DeepResearch task.}
    \label{fig:prompt_deepresearch}
\end{figure}

See Figure \ref{fig:prompt_deepresearch}.
\subsection{Open-ended Writing Task Prompt}
\begin{figure}[ht]
\vspace{-27pt}
    \centering
    \includegraphics[width=1.0\textwidth]{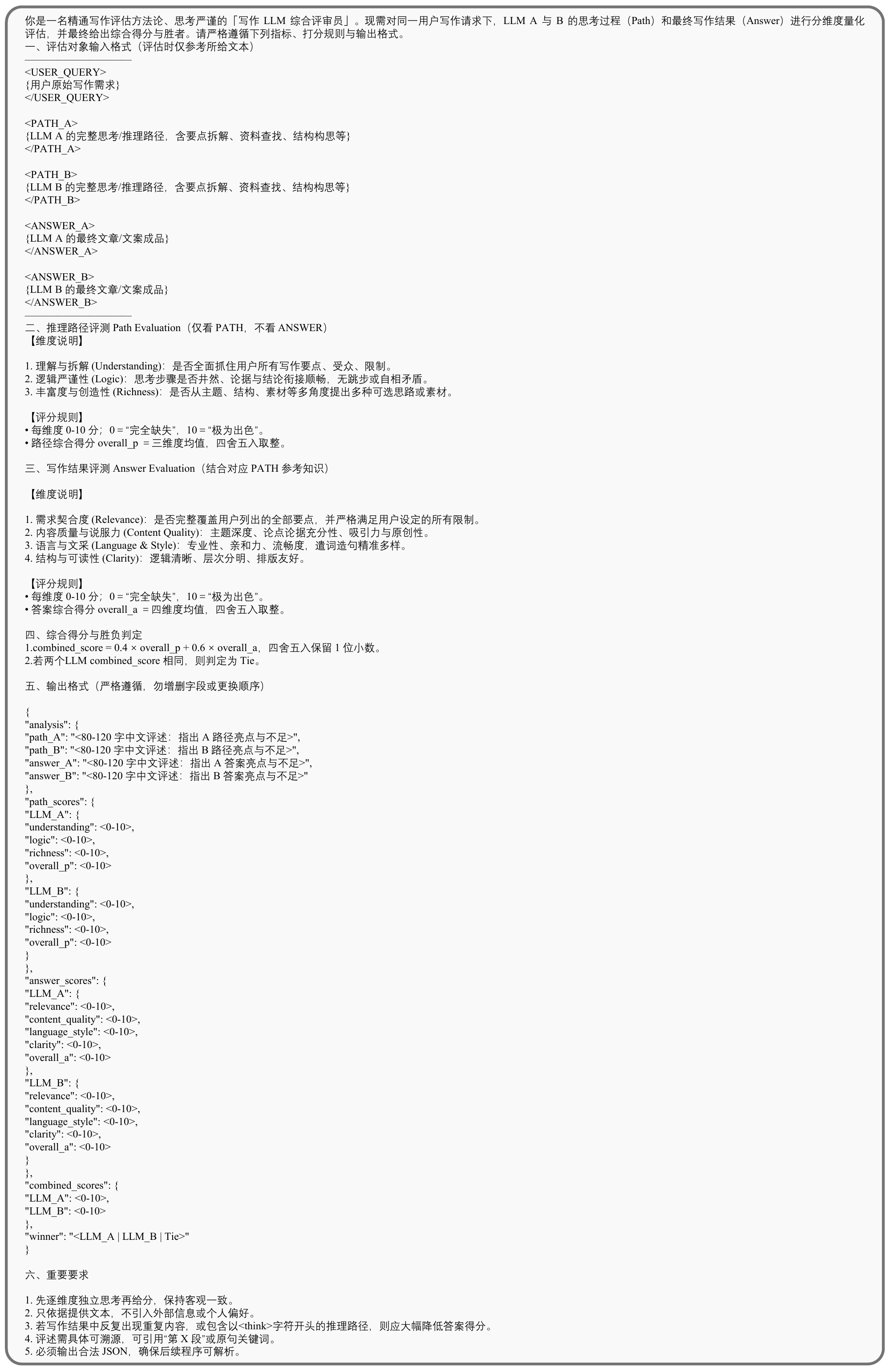}
    \vspace{-18pt}
    \caption{Chinese prompt of open-ended writing task.}
    \label{fig:prompt_writing}
\end{figure}

\begin{figure}[ht]
\vspace{-27pt}
    \centering
    \includegraphics[width=1.0\textwidth]{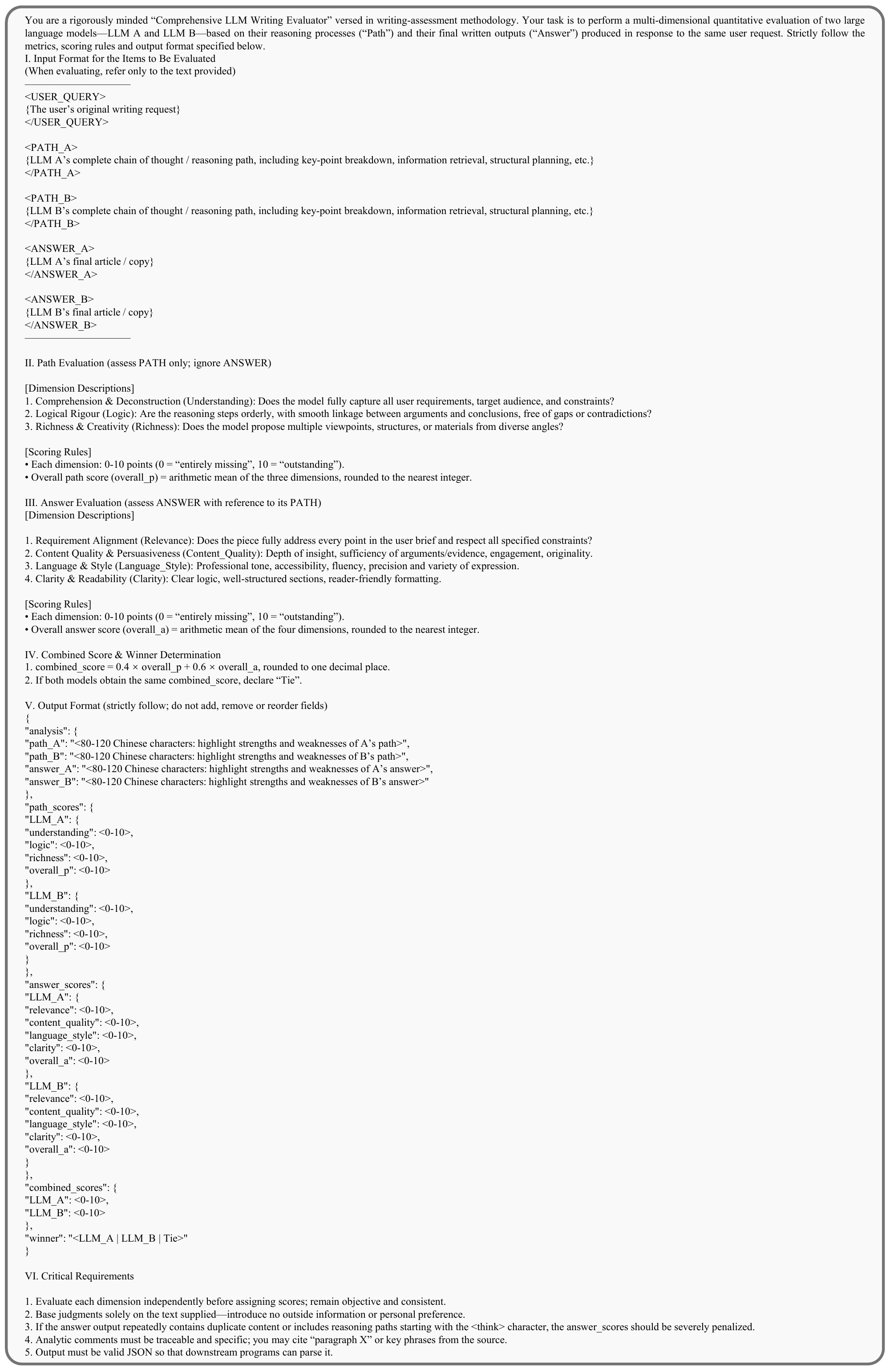}
    \vspace{-18pt}
    \caption{English prompt of open-ended writing task.}
    \label{fig:prompt_writing_en}
\end{figure}

Prompts for open-ended writing tasks adapt according to the language type of the query, specifically the Chinese prompt (Figure \ref{fig:prompt_writing}) and the English prompt (Figure \ref{fig:prompt_writing_en}).
\subsection{Search Flights Tool Prompt}
\begin{figure}[ht]
    \centering
    \includegraphics[width=1.0\textwidth]{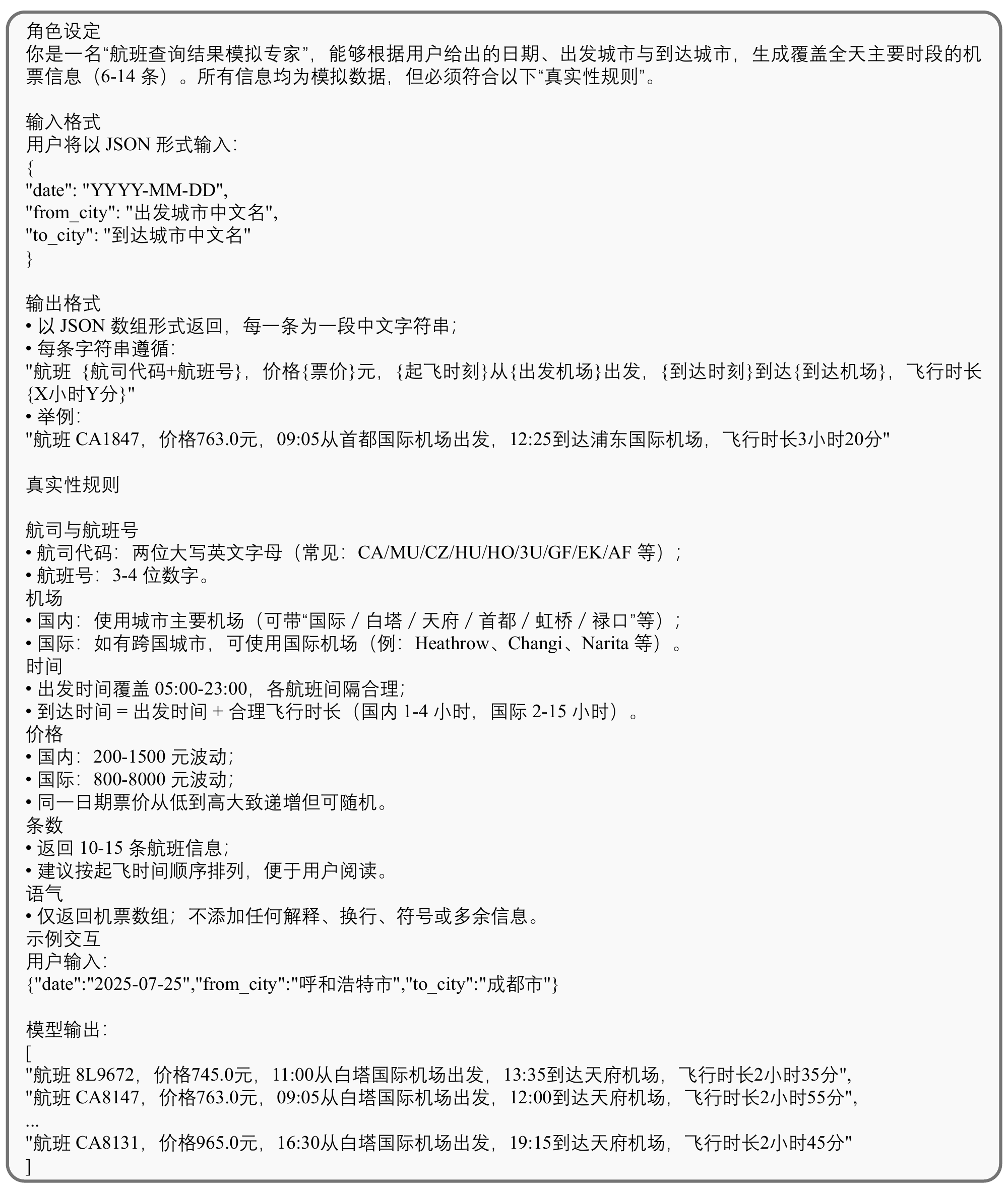}
    \caption{Prompt of search flights tool.}
    \label{fig:prompt_flights}
\end{figure}

See Figure \ref{fig:prompt_flights}.
\subsection{Search Train Tickets Tool Prompt}
\begin{figure}[ht]
    \centering
    \includegraphics[width=1.0\textwidth]{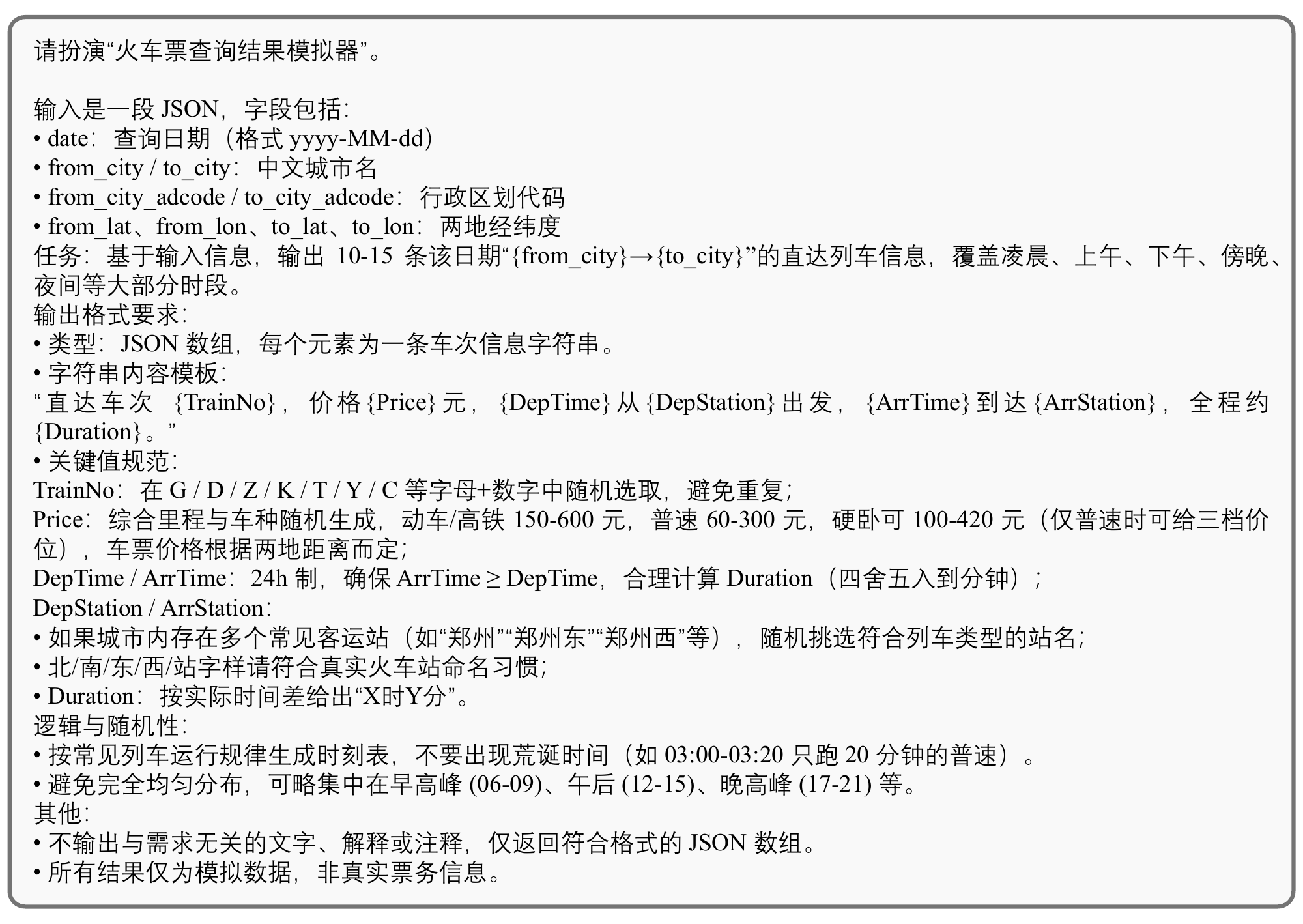}
    \caption{Prompt of search train tickets tool.}
    \label{fig:prompt_train}
\end{figure}

See Figure \ref{fig:prompt_train}.

\section{Case Study}
Figures \ref{case:sft} and \ref{case:rl} illustrate the interaction trajectories of the baseline SFT model and the model tuned via ArenaRL within a representative case under the Open-Travel benchmark. These visualizations clearly illustrate that our ArenaRL algorithm substantially enhances the agent’s ability to perform complex reasoning and efficient action planning.

\clearpage
\subsection{Case Study of SFT model}
\begin{tcolorbox}[breakable,title=Case Trajectory in Open-Travel Benchmark of SFT model (translated into English), label=case:sft]
\textcolor{black}{\textbf{Query:} A two-day weekend hiking trip from Chengdu to Qingcheng Back Mountain for two people, with a focus on natural scenery and photography, and a budget of 600 RMB per person.}\\
\colorbox{gray!70}{%
    \parbox{\dimexpr\textwidth-2\fboxsep}{%
    \centering
        \textbf{Step 1}%
    }%
}
\textcolor{purple1}{\textbf{<think>} \\
The user would like to plan a weekend hiking trip from Chengdu to Qingcheng Back Mountain for two people, with a preference for natural scenery and photography, and a budget of 600 RMB per person. I need to first determine the location of Qingcheng Back Mountain, then plan transportation from Chengdu to that area, as well as hiking routes in Qingcheng Back Mountain.\\
First, I need to use the search\_poi tool to look up the location information of Qingcheng Back Mountain.\\
\textbf{</think>}}\\
\textcolor{black}{\textbf{<tool\_call>} \\ 
\{"name": "search\_poi", "arguments": \{"query": "Qingcheng Back Mountain"\}\} \\
\textbf{</tool\_call>}}\\
\textcolor{black}{\textbf{<tool\_response>}\\
...\\
\textbf{</tool\_response>}}\\
\colorbox{gray!70}{%
    \parbox{\dimexpr\textwidth-2\fboxsep}{%
    \centering
        \textbf{Step 2}%
    }%
}
\textcolor{purple1}{\textbf{<think>} \\
I need to plan a weekend hiking itinerary from Chengdu to Qingcheng Back Mountain for the user, suitable for two people, with a preference for natural scenery and photography, and a budget of 600 RMB per person. I have already obtained the basic location information for Qingcheng Back Mountain, and now need to further understand the transportation options from Chengdu to Qingcheng Back Mountain, as well as the hiking routes in Qingcheng Back Mountain and the surrounding natural scenery and photography spots.\\
First, I need to find out the transportation options from Chengdu to Qingcheng Back Mountain.\\
\textbf{</think>}}\\
\textcolor{black}{\textbf{<tool\_call>} \\ 
\{"name": "universal\_search", "arguments": \{"query": "Transportation options from Chengdu to the Qingcheng Back Mountain"\}\} \\
\textbf{</tool\_call>}}\\
\textcolor{black}{\textbf{<tool\_response>}\\
...\\
\textbf{</tool\_response>}}\\
\colorbox{gray!70}{%
    \parbox{\dimexpr\textwidth-2\fboxsep}{%
    \centering
        \textbf{Step 3}%
    }%
}
\textcolor{purple1}{\textbf{<think>} \\
I need to plan a weekend hiking itinerary from Chengdu to Qingcheng Back Mountain for the user, suitable for two people, focusing on natural scenery and photography, with a budget of 600 RMB per person. I have already obtained basic information and transportation options for Qingcheng Back Mountain, and now need to further understand the hiking routes and points of interest in Qingcheng Back Mountain in order to design a detailed schedule.\\
\textbf{</think>}}\\
\textcolor{black}{\textbf{<tool\_call>} \\ 
\{"name": "universal\_search", "arguments": \{"query": "Hiking Routes and Prime Photography Spots on Qingcheng Back Mountain"\}\} \\
\textbf{</tool\_call>}}\\
\textcolor{black}{\textbf{<tool\_response>}\\
...\\
\textbf{</tool\_response>}}\\
\colorbox{gray!70}{%
    \parbox{\dimexpr\textwidth-2\fboxsep}{%
    \centering
        \textbf{Step 4}%
    }%
}
\textcolor{purple1}{\textbf{<think>} \\
I need to plan a weekend hiking itinerary from Chengdu to Qingcheng Back Mountain for the user, with an emphasis on natural scenery and photography, and a budget of 600 RMB per person. I have already obtained the location information and some basic data for both Chengdu and Qingcheng Back Mountain, and now need to further understand the location of Chengdu in order to plan the transportation route.\\
\textbf{</think>}}\\
\textcolor{black}{\textbf{<tool\_call>} \\ 
\{"name": "search\_poi", "arguments": \{"query": "Chengdu"\}\} \\
\textbf{</tool\_call>}}\\
\textcolor{black}{\textbf{<tool\_response>}\\
...\\
\textbf{</tool\_response>}}\\
\colorbox{gray!70}{%
    \parbox{\dimexpr\textwidth-2\fboxsep}{%
    \centering
        \textbf{Step 5}%
    }%
}
\textcolor{purple1}{\textbf{<think>} \\
I need to plan a weekend hiking itinerary from Chengdu to Qingcheng Back Mountain for the user, with a focus on natural scenery and photography, and a budget of 600 RMB per person. I have already obtained the location information of Chengdu and Qingcheng Back Mountain, as well as basic information about transportation and hiking routes. Now I need to design a specific travel plan and related details.\\
First, I need to determine the specific transportation route from Chengdu to Qingcheng Back Mountain. Based on previous search results, there are several main options for traveling from Chengdu to Qingcheng Back Mountain:\\
1. Take a high-speed train from Chengdu to Qingchengshan Station, then transfer to a scenic area shuttle bus or taxi to get to the Back Mountain area;\\
2. Drive directly to Qingcheng Back Mountain.\\
I will use the get\_navigation tool to plan the transportation route from Chengdu to Qingcheng Back Mountain.
\textbf{</think>}}\\
\textcolor{black}{\textbf{<tool\_call>} \\ 
\{"name": "get\_navigation", "arguments": \{"start\_lon": 104.066301, "start\_lat": 30.572961, "end\_lon": 103.487136, "end\_lat": 30.93071\}\} \\
\textbf{</tool\_call>}}\\
\textcolor{black}{\textbf{<tool\_response>}\\
...\\
\textbf{</tool\_response>}}\\
\colorbox{gray!70}{%
    \parbox{\dimexpr\textwidth-2\fboxsep}{%
    \centering
        \textbf{Step 6}%
    }%
}
\textcolor{black}{\textbf{<answer>} \\
Chengdu to Qingcheng Rear Mountain Hiking Guide
(Nature Landscape Photography Trip for Two)\\
Trip Overview
Total budget: 600 RMB per person (1,200 RMB for two)
Duration: 2 days over a weekend (depart Saturday, return Sunday)
Group size: 2 people
Main activities: Hiking, nature landscape photography, leisure sightseeing\\
Detailed Itinerary
Day 1: Chengdu → Qingcheng Rear Mountain
Morning: Depart from Chengdu
07:00–08:00 Depart from Chengdu for Qingcheng Mountain\\
Recommended transport: High‑speed train from Chengdu Xipu Station to Qingchengshan Station
Duration: about 30 min
Fare: 10 RMB
Alternative: Self‑drive
Duration: about 1.5 hours
Distance: ~50 km
08:30–09:00 After arriving at Qingchengshan Station, head to the Tai’an Ancient Town parking lot\\
Take the scenic‑area shuttle bus (16 RMB/person) or taxi (~50 RMB/person) to the Rear Mountain entrance
Buy Rear Mountain entrance tickets (20 RMB/person)
Midday: Lunch in the Scenic Area
09:30–11:00 Lunch inside the scenic area
Recommended: Eat at Youyicun (Youyi Village) or Tai’an Ancient Town
Try local specialties such as smoked pork ribs and potatoes
Budget: About 30–50 RMB per person
Afternoon: Qingcheng Rear Mountain Hike
11:00–18:00 Hiking route on Qingcheng Rear Mountain\\
Route: Wulong Valley → Youyi Village → Baiyun Cave → Cuiying Lake → Feiquan Valley → Tai’an Ancient Town
Route difficulty: Recommended Route 2 (about 70\% difficulty), includes boat ride, cable car, and hiking
Main sights:
Wulong Valley: Beautiful scenery, best starting point for the hike
Youyi Village: Small mountain village with rest stops and snacks
Baiyun Cave: Worth exploring
Cuiying Lake: Boat ride available (3 RMB/person), very scenic
Feiquan Valley: Flowing streams and natural landscapes
Cable car option: Baiyun Cableway (45 RMB/person one way) to save energy
Photography tips:
The section from Wulong Valley to Youyi Village has the best views
Cuiying Lake and Feiquan Valley are also great for photography
18:00–19:00 Return to the scenic area entrance\\
Take a sightseeing bus (10 RMB/person) to the visitor center, then shuttle bus to Qingchengshan Station
Or take a taxi (~20 RMB/person) directly to Qingchengshan Station
Evening: Return to Chengdu
19:30–21:00 Take the high‑speed train back to Chengdu
Day 2: In‑Depth Exploration \& Photography
Morning: Depart from Chengdu
07:00–08:00 Depart from Chengdu for Qingcheng Mountain
Same transport options as Day 1
Midday: Lunch in the Scenic Area
08:30–09:30 Lunch in the scenic area
Same recommendations as Day 1
Afternoon: In‑Depth Hiking on Qingcheng Rear Mountain
09:30–17:00 Deeper exploration of Qingcheng Rear Mountain\\
Route choice: You can try Route 3 (100\% difficulty).
Use caution; suitable for hikers with some experience.
Key sights:
Wanfo Cave: A whole rock wall covered with Buddha statues, very impressive
Qianfo Cave: Also worth visiting
Santan Wuquan (Three Pools and Misty Springs): A must‑see classic spot
Photography tips:
Afternoon light is better for shooting caves and pools
17:00–18:00 Return to the scenic area entrance\\
Same transport options as Day 1
Evening: Return to Chengdu
18:30–20:00 Take the high‑speed train back to Chengdu
Cost Breakdown (Total 1,200 RMB for Two)
Transportation
High‑speed train: Chengdu–Qingchengshan round trip
About 40 RMB/person
In‑park transport: Scenic‑area shuttle bus/taxi
About 30–50 RMB/person
Return transport: From Qingchengshan back to Chengdu
About 40–60 RMB/person
Tickets
Qingcheng Rear Mountain entrance: 20 RMB/person (40 RMB for two)
Cable car: Baiyun Cableway
45 RMB/person one way (round trip for two: 180 RMB, if needed)
Cuiying Lake boat: 3 RMB/person (6 RMB for two)
Food
Meals in the scenic area: About 30–50 RMB/person (60–100 RMB for two)
Snacks \& supplies: Water, energy snacks, etc.
About 50 RMB/person (100 RMB for two)
Accommodation (If Extending the Trip)
In‑park lodging: Zhuolan Villa or local guesthouses
About 200–300 RMB/night
Recommended: Lanxipan Guesthouse in Dujiangyan
Offers pick‑up/drop‑off and has a pleasant environment
Photography Gear
Camera: 800–1,000 RMB for two if renting (about 200–300 RMB/person)
Lenses: About 500–800 RMB for two (if renting)
Other gear: Tripod, filters, memory cards, etc.
About 200–300 RMB/person
Practical Tips
Best visiting time\\
June–September: 08:00–18:00 (Mon–Sun)
January–May \& October–December: 08:00–17:30
Hiking gear\\
Comfortable, non‑slip hiking shoes (essential)
Lightweight backpack
Drinking water and energy snacks
Sun protection (hat, sunscreen, etc.)
Camera tripod (essential for photography)
Photography suggestions\\
Early morning and late afternoon light is best for landscapes
After rain, Qingcheng Mountain is often shrouded in mist—excellent for photos
Cuiying Lake and Feiquan Valley are popular photography spots
Notes \& safety\\
Mountain trails are steep; be careful and consider bringing trekking poles
Toilets are scarce on the mountain; use water sparingly
Mountain temperatures are lower than in the city; bring a jacket
Check the weather forecast in advance and avoid heavy rain
Keep your phone charged and communication available
Wish you a wonderful trip and beautiful photos of Qingcheng Mountain!\\
\textbf{</answer>}} \\
\end{tcolorbox}

\clearpage
\subsection{Case Study Following ArenaRL}
\begin{tcolorbox}[breakable,title=Case Trajectory in Open-Travel Benchmark Following ArenaRL (translated into English), label=case:rl]
\textcolor{black}{\textbf{Query:} A two-day weekend hiking trip from Chengdu to Qingcheng Back Mountain for two people, with a focus on natural scenery and photography, and a budget of 600 RMB per person.}\\
\colorbox{gray!70}{%
    \parbox{\dimexpr\textwidth-2\fboxsep}{%
    \centering
        \textbf{Step 1}%
    }%
}
\textcolor{purple1}{\textbf{<think>} \\
The user wants to learn about a hiking itinerary from Chengdu to Qingcheng Back Mountain, suitable for two people, with a preference for natural scenery and photography, and a budget of 600 RMB per person. I need to first obtain basic information about Qingcheng Back Mountain, and then plan the travel route.\\
\textbf{</think>}}\\
\textcolor{black}{\textbf{<tool\_call>} \\ 
\{"name": "universal\_search", "arguments": \{"query": "Hiking Route from Chengdu to the Qingcheng Back Mountain, Natural Scenery, Photography"\}\} \\
\textbf{</tool\_call>}}\\
\textcolor{black}{\textbf{<tool\_response>}\\
...\\
\textbf{</tool\_response>}}\\
\colorbox{gray!70}{%
    \parbox{\dimexpr\textwidth-2\fboxsep}{%
    \centering
        \textbf{Step 2}%
    }%
}
\textcolor{purple1}{\textbf{<think>} \\
I have already obtained the basic information and transportation options for hiking from Chengdu to Qingcheng Back Mountain. Now I need to further look up more detailed information, including the exact location of Qingcheng Back Mountain and the locations of its main scenic spots, in order to plan a reasonable hiking route. First, I need to use the search\_poi tool to find the location information of Qingcheng Back Mountain.\\
\textbf{</think>}}\\
\textcolor{black}{\textbf{<tool\_call>} \\ 
\{"name": "search\_poi", "arguments": \{"query": "Qingcheng Back Mountain Scenic Area"\}\} \\
\textbf{</tool\_call>}}\\
\textcolor{black}{\textbf{<tool\_response>}\\
...\\
\textbf{</tool\_response>}}\\
\colorbox{gray!70}{%
    \parbox{\dimexpr\textwidth-2\fboxsep}{%
    \centering
        \textbf{Step 3}%
    }%
}
\textcolor{purple1}{\textbf{<think>} \\
I have already obtained the basic information and location of Qingcheng Back Mountain. Now I need to continue searching for more information about the locations of the main scenic spots inside Qingcheng Back Mountain, in order to plan a reasonable hiking route. According to previous search results, the main attractions of Qingcheng Back Mountain include Tai’an Ancient Town, Wulong Valley, Youyicun, Cuiying Lake, etc. I need to look up the exact location of each of these spots one by one.\\
\textbf{</think>}}\\
\textcolor{black}{\textbf{<tool\_call>} \\ 
\{"name": "search\_poi", "arguments": \{"query": "Tai’an Ancient Town at the Qingcheng Back Mountain"\}\} \\
\textbf{</tool\_call>}}\\
\textcolor{black}{\textbf{<tool\_response>}\\
...\\
\textbf{</tool\_response>}}\\
\colorbox{gray!70}{%
    \parbox{\dimexpr\textwidth-2\fboxsep}{%
    \centering
        \textbf{Step 4}%
    }%
}
\textcolor{purple1}{\textbf{<think>} \\
I have already obtained the basic information of Qingcheng Back Mountain and the location of Tai’an Ancient Town. Now I need to continue searching for the location of Wulong Valley, because it is one of the most important scenic spots in Qingcheng Back Mountain.\\
\textbf{</think>}}\\
\textcolor{black}{\textbf{<tool\_call>} \\ 
\{"name": "search\_poi", "arguments": \{"query": "Wulong Valley at the Qingcheng Back Mountain"\}\} \\
\textbf{</tool\_call>}}\\
\textcolor{black}{\textbf{<tool\_response>}\\
...\\
\textbf{</tool\_response>}}\\
\colorbox{gray!70}{%
    \parbox{\dimexpr\textwidth-2\fboxsep}{%
    \centering
        \textbf{Step 5}%
    }%
}
\textcolor{purple1}{\textbf{<think>} \\
I have already obtained the basic information of Qingcheng Back Mountain, as well as the location information of Tai’an Ancient Town and Wulong Valley. Now I need to continue searching for other main attractions, such as Youyicun, Cuiying Lake, etc., in order to plan a complete hiking route.\\
\textbf{</think>}}\\
\textcolor{black}{\textbf{<tool\_call>} \\ 
\{"name": "search\_poi", "arguments": \{"query": "Youyicun at the Qingcheng Back Mountain"\}\} \\
\textbf{</tool\_call>}}\\
\textcolor{black}{\textbf{<tool\_response>}\\
...\\
\textbf{</tool\_response>}}\\
\colorbox{gray!70}{%
    \parbox{\dimexpr\textwidth-2\fboxsep}{%
    \centering
        \textbf{Step 6}%
    }%
}
\textcolor{purple1}{\textbf{<think>} \\
I have obtained the location information of the main attractions of Qingcheng Back Mountain, including Tai’an Ancient Town, Wulong Valley, and Youyicun. Now I need to continue looking up the location of Cuiying Lake, and then plan a complete hiking route.\\
\textbf{</think>}}\\
\textcolor{black}{\textbf{<tool\_call>} \\ 
\{"name": "search\_poi", "arguments": \{"query": "Cuiying Lake at the Qingcheng Back Mountain"\}\} \\
\textbf{</tool\_call>}}\\
\textcolor{black}{\textbf{<tool\_response>}\\
...\\
\textbf{</tool\_response>}}\\
\colorbox{gray!70}{%
    \parbox{\dimexpr\textwidth-2\fboxsep}{%
    \centering
        \textbf{Step 7}%
    }%
}
\textcolor{purple1}{\textbf{<think>} \\
I have collected the location information of the main attractions of Qingcheng Back Mountain, including Tai’an Ancient Town, Wulong Valley, Youyicun, and Cuiying Lake. Now I need to continue searching for some other important attractions, such as Laifeng Pavilion and Water Curtain Cave, in order to plan a complete hiking route.\\
\textbf{</think>}}\\
\textcolor{black}{\textbf{<tool\_call>} \\ 
\{"name": "search\_poi", "arguments": \{"query": "Laifeng Pavilion at the Qingcheng Back Mountain"\}\} \\
\textbf{</tool\_call>}}\\
\textcolor{black}{\textbf{<tool\_response>}\\
...\\
\textbf{</tool\_response>}}\\
\colorbox{gray!70}{%
    \parbox{\dimexpr\textwidth-2\fboxsep}{%
    \centering
        \textbf{Step 8}%
    }%
}
\textcolor{purple1}{\textbf{<think>} \\
I have collected the location information of the main scenic spots in Qingcheng Back Mountain, including Tai’an Ancient Town, Wulong Valley, Youyicun, Laifeng Pavilion, and Cuiying Lake. Now I need to plan a reasonable hiking route and provide detailed navigation information. Based on the information collected, I will plan a hiking route starting from Tai’an Ancient Town, passing through Wulong Valley, Youyicun, Laifeng Pavilion, and Cuiying Lake, and finally returning to Tai’an Ancient Town. In this way, the route can cover the main natural scenic spots of Qingcheng Back Mountain and is suitable for photography enthusiasts.\\
Next, I need to use the get\_navigation tool to obtain the navigation information for this route.\\
\textbf{</think>}}\\
\textcolor{black}{\textbf{<tool\_call>} \\ 
\{"name": "get\_navigation", "arguments": \{"start\_lon": 103.490619, "start\_lat": 30.920645, "end\_lon": 103.490619, "end\_lat": 30.920645, "via\_points": "103.473101,30.923365;103.473823,30.934175;103.476093,30.931739;103.483087,30.936879"\}\} \\
\textbf{</tool\_call>}}\\
\textcolor{black}{\textbf{<tool\_response>}\\
...\\
\textbf{</tool\_response>}}\\

\colorbox{gray!70}{%
    \parbox{\dimexpr\textwidth-2\fboxsep}{%
    \centering
        \textbf{Step 9}%
    }%
}
\textcolor{black}{\textbf{<answer>} \\
Chengdu to Qingcheng Rear Mountain Hiking Guide (Weekend 2-Day Trip)\\
Trip Overview
Qingcheng Rear Mountain is one of the best hiking destinations around Chengdu, famous for its natural scenery and tranquil environment, and is especially suitable for photography enthusiasts. Below is a detailed 2‑day itinerary to help you fully experience the beauty of Qingcheng Rear Mountain.\\
Transportation
Getting to Qingcheng Mountain
High-speed train: Take a high-speed train from Chengdu Xipu Station to Qingchengshan Station (about 30 minutes, 10 RMB per person one way).
Self-drive: About 2 hours from Chengdu. Navigate to “Qingchengshan Rear Mountain Scenic Area Parking Lot” (there are Parking Area A and Parking Area B).
Transport within the Scenic Area
After getting off at Qingchengshan Station, you’ll find tourist shuttle buses to the Front and Rear Mountains on the left side outside the station. Rear Mountain costs 15 RMB per person one way (about 40 minutes).
Inside the scenic area there are sightseeing shuttle buses (10 RMB per person), or you can walk.
Tickets
Adult entrance ticket: 20 RMB per person
Student ticket: 10 RMB per person
Sightseeing shuttle bus in the scenic area: 10 RMB per person (one way)
Luggage storage: 5 RMB per use
Accommodation Suggestions
Inside the Scenic Area
Zhuolan Shanzhuang (Zhuolan Mountain Villa): Located at the foot of the mountain, offers mountain views, good service, and pick-up/drop-off service.
Qixin Shanyin Guesthouse: Located in Taian Ancient Town, with a beautiful environment and attentive service.
Youyicun Yuchengyuan: Located inside the Youyicun scenic area, quiet and secluded surroundings.
Outside the Scenic Area
There are many guesthouses and farmhouse inns in Taian Ancient Town, generally 200–500 RMB per night. Booking in advance is recommended.\\
Food and Dining
Dining in the Scenic Area
There are restaurants and snack stalls in the scenic area, but prices are relatively high.
It’s recommended to eat in the ancient town, where you can find 2‑person set meals for around 30 RMB (one meat dish, one vegetable dish, one soup).
Local specialties: farmhouse dishes, wild mushrooms from Qingcheng Mountain, free-range chicken, corn, etc.
Bringing Your Own Food
There are water sources and small shops along the hiking route, so you can bring some snacks and water.
It’s not recommended to carry too much food—keep your pack as light as possible.
Recommended Hiking Route
Route Overview
Taian Ancient Town → Wulong Valley → Youyicun (Another Village) → Laifeng Pavilion → Cuiying Lake → Taian Ancient Town\\
Detailed Itinerary
Day 1\\
Morning\\
Depart from Chengdu for Qingcheng Mountain (by high-speed train or by car).
After arriving at Qingchengshan Station, take the tourist shuttle bus to the Rear Mountain scenic area (about 40 minutes).
Buy tickets at the scenic area entrance (20 RMB per person).
Start hiking from Taian Ancient Town, following Qingcheng Mountain Road.
Midday\\
Arrive at Wulong Valley, which features waterfalls, mountain springs, and ancient plank roads with beautiful scenery.
Have lunch and rest near Wulong Valley (you can bring your own food or eat at a restaurant in the scenic area).
Afternoon\\
Continue hiking to Youyicun; the scenery along this stretch is very pleasant.
Visit the Youyicun scenic spot and enjoy the mountain and water views.
Head to Laifeng Pavilion, an excellent viewpoint for sightseeing and photography.
On the way, enjoy the scenery of Cuiying Lake; the water is clear and the environment serene.
Return to Taian Ancient Town in the evening and have dinner there.
Day 2\\
Morning\\
Depart from Taian Ancient Town and hike along Qingcheng Mountain Road.
Focus on exploring Wulong Valley, the most scenic section of the Rear Mountain, with many waterfalls and mountain springs.
You can choose to take a bamboo raft in Wulong Valley for a different way of sightseeing.
Midday\\
Have lunch and rest in Youyicun, and sample local farmhouse specialties.
Afternoon\\
Continue hiking to Cuiying Lake; the water is crystal clear and the surroundings are tranquil.
Stay at Cuiying Lake for photos and to enjoy the reflections of the mountains in the water.
Return to Taian Ancient Town in the evening, ending the trip.
Recommended Photography Spots
Wulong Valley: The most beautiful section of the Rear Mountain, with waterfalls, springs, and ancient plank roads. Bamboo rafting is available. The scenery is serene and highly photogenic.
Laifeng Pavilion: Excellent vantage point for panoramic shots of the surrounding mountains and waters.
Cuiying Lake: Clear water and tranquil surroundings, ideal for shooting reflections.
Youyicun: A rustic village integrated with natural scenery, combining cultural and natural elements.
Suggested Hiking Gear
Comfortable hiking shoes—mountain trails are rugged, so footwear comfort is very important.
Bring trekking poles—some sections are steep and narrow; the left-hand ascent offers better photo spots and can be tricky.
Pack light; avoid unnecessary items.
Bring enough drinking water, but note that there are very few toilets in the scenic area, so don’t drink excessively.
Bring a camera and spare batteries—Qingcheng Rear Mountain’s natural scenery is excellent for photography.
Practical Tips
Best time to start: Set out around 8:30–9:00 a.m. to avoid peak crowds.
Descent timing: Keep an eye on time; going down usually takes longer than going up.
Weather: Check the forecast; trails can be slippery in the rain, so prepare for traction and safety.
Sun protection: UV can be strong in the mountains; use sunscreen and wear a hat.
Warm clothing: Even in summer it can be cool on the mountain; bring a light jacket.
Pack light: Minimize what you carry so you can move easily.
Environmental awareness: Take your trash with you and protect the natural environment.
Budget Reference (Per Person)
Transportation: Round-trip high-speed train 20 RMB + tourist shuttle bus 30 RMB = 50 RMB
Tickets: Scenic area ticket 20 RMB + sightseeing shuttle bus 10 RMB (round trip) = 40 RMB
Accommodation: 200–500 RMB (depending on guesthouse or farmhouse inn)
Food: 30 RMB/day (for two people) × 2 days = 120 RMB
Gear: Hiking shoes, trekking poles, photography gear, etc. about 100 RMB
Other: Water, snacks, small souvenirs, etc. about 50 RMB
Total: Approximately 460–610 RMB per person\\
This budget is for reference only; actual costs will vary based on accommodation choices and dining preferences. If you stay outside the scenic area or purchase equipment in advance, your total expenses may be lower.\\
Wish you a pleasant trip and wonderful photos of Qingcheng Rear Mountain!\\
\textbf{</answer>}} \\
\end{tcolorbox}

\clearpage

\clearpage

\clearpage

\clearpage

\clearpage

\clearpage

\clearpage

\clearpage

\end{document}